\documentclass[Preprint,12pt,authoryear]{elsarticle}
\usepackage{graphicx,amssymb,mathrsfs,amsmath,pifont,amscd,latexsym,amsfonts,amsthm,colortbl}
\usepackage{geometry}
\usepackage{graphics}
\usepackage{epstopdf}
\usepackage{float}
\usepackage{subfigure}
\usepackage{epsfig}
\usepackage{setspace}
\usepackage{booktabs}
\usepackage{xcolor}
\usepackage{multirow}
\usepackage{soul}
\usepackage{algorithm}
\usepackage{algpseudocode}
\usepackage{cite}
\usepackage[colorlinks,urlcolor=blue,driverfallback=dvipdfm]{hyperref}
\journal{ISPRS Journal of Photogrammetry and Remote Sensing}

\newcommand{\norm}[1]{\lVert#1\rVert}
\newcommand{\cmark}{\ding{51}}
\newcommand{\xmark}{\ding{55}}

\DeclareMathOperator{\F}{F}

\DeclareMathOperator{\tr}{tr}
\DeclareMathOperator{\svd}{svd}

\begin{document}

\begin{frontmatter}
\title{Multimodal Remote Sensing Benchmark Datasets for Land Cover Classification with A Shared and Specific Feature Learning Model}

\author{
Danfeng Hong\textsuperscript{a}, Jingliang Hu\textsuperscript{b}, Jing Yao\textsuperscript{c}, \\ Jocelyn Chanussot\textsuperscript{e,c}, Xiao Xiang Zhu\textsuperscript{a,b,*}}

\address{
	\textsuperscript{a}Remote Sensing Technology Institute, German Aerospace Center, 82234 Wessling, Germany\\
	\textsuperscript{b}Data Science in Earth Observation, Technical University of Munich, 80333 Munich, Germany\\
	\textsuperscript{c}Aerospace Information Research Institute, Chinese Academy of Sciences, 100094 Beijing, China\\
	\textsuperscript{e}Univ. Grenoble Alpes, INRIA, CNRS, Grenoble INP, LJK, 38000 Grenoble, France\\
	*Corresponding author\\
}
\begin{abstract}
\textcolor{blue}{This is the pre-acceptance version, to read the final version please go to ISPRS Journal of Photogrammetry and Remote Sensing.}As remote sensing (RS) data obtained from different sensors become available largely and openly, multimodal data processing and analysis techniques have been garnering increasing interest in the RS and geoscience community. However, due to the gap between different modalities in terms of imaging sensors, resolutions, and contents, embedding their complementary information into a consistent, compact, accurate, and discriminative representation, to a great extent, remains challenging. To this end, we propose a shared and specific feature learning (S2FL) model. S2FL is capable of decomposing multimodal RS data into modality-shared and modality-specific components, enabling the information blending of multi-modalities more effectively, particularly for heterogeneous data sources. Moreover, to better assess multimodal baselines and the newly-proposed S2FL model, three multimodal RS benchmark datasets, i.e., \textit{Houston2013} -- hyperspectral and multispectral data, \textit{Berlin} -- hyperspectral and synthetic aperture radar (SAR) data, \textit{Augsburg} -- hyperspectral, SAR, and digital surface model (DSM) data, are released and used for land cover classification. Extensive experiments conducted on the three datasets demonstrate the superiority and advancement of our S2FL model in the task of land cover classification in comparison with previously-proposed state-of-the-art baselines. Furthermore, the baseline codes and datasets used in this paper will be made available freely at \url{https://github.com/danfenghong/ISPRS_S2FL}.
\end{abstract}

\end{frontmatter}

\graphicspath{{Figures/}}

\section{Introduction}
The rapid development of remotely sensed imaging techniques enables the measurement and monitoring of Earth on the land surface and beneath (e.g., identification of underground minerals \citep{bishop2011hyperspectral}, geological environment survey and monitoring \citep{van2012multi}, volcanic terrain component analysis \citep{amici2013geological}), of the quality of air and water, and of the health of humans, plants, and animals \citep{nativi2015big}. Remote sensing (RS) is one of the most important contact-free sensing means for Earth observation (EO) to extract relevant information about the physical properties of the Earth and environment system from spaceborne and airborne platforms. With the ever-growing availability of RS data sources from both satellite and airborne sensors on a large scale and even global scale, multimodal RS image processing and analysis techniques have been garnering growing attention in various EO-related tasks\citep{schmitt2016}, such as land cover change detection \citep{liu2017band,liu2019review}, disaster monitoring and management \citep{zhu2019integrating,liu2020novel}, urban planning \citep{weng2009thermal,xie2017spatiotemporally}, mineral exploration \citep{hong2019augmented,siebels2020estimation}. 

The data acquired by different platforms can provide diverse and complementary information, including light detection and ranging (LiDAR) or digital surface model (DSM) providing the height information about the ground elevation, synthetic aperture radar (SAR) providing the structure information about Earth’s surface, and multispectral (MS) or hyperspectral (HS) data providing detailed content information of sensed materials. The joint exploitation of different RS data has been therefore proven to be useful to further enhance the understanding, possibilities, and capabilities to Earth and our environment. In a complex urban scene, the ability of spectral data (e.g., RGB, MS, HS) in finely identifying the land cover categories usually remains limited, particularly for those categories that have extremely similar spatial structure or spectral signatures. For example, the material ``Asphalt'' on the road or on the roof can be hardly classified by only observing subtle discrepancies from their spectral profiles \citep{heiden2012urban}. But fortunately, we might expect to have the height information provided by DSM data, which enables the fine-grained recognition of these similar materials at a higher accuracy compared to only single modalities. It is well known that HS images are characterized by nearly continuous spectral properties, while MS images can provide finer spatial information. The fusion of MS and HS images naturally becomes a feasible solution to obtain the image product of high spatial and spectral resolutions. Another common example is cloud removal. Optical RS data (e.g., RGB, MS, HS) tend to suffer from the cloud occlusion in imaging process, thereby bringing the risk of important information loss. SAR data or DSM data generated from the LiDAR are insensitive to the cloud coverage by capturing intrinsic structure or elevation information that are not closely associated with the cloud \citep{gao2020cloud}.

In recent years, many previous works have been proposed by the attempts to boost the development of multimodal RS techniques in land cover classification. Nevertheless, the ability of these approaches in multimodal RS data representations remains limited, further limiting the performance gain of subsequent high-level applications. This can be well explained by two possible factors as follows.
\begin{itemize}
    \item {\small \textbf{Multimodal RS Benchmark Datasets.}} Despite the increased availability of the multimodal RS data, different contexts, structures, sensors, resolutions, and imaging conditions pose a great challenge in data acquisition and processing for to-be-studied scenes \citep{dalla2015challenges}. Consequently, the lack of multimodal RS benchmark datasets, to a larger extent, limits the development of the corresponding methodologies and the practical application of land cover classification.
    \item {\small \textbf{Multimodal Feature Learning Models.}} Feature extraction (FE), as one of the most important steps prior to the high-level data analysis, also plays a key role in multimodal RS. Most of previous FE methods, either hand-crafted or learning-based \citep{rasti2020feature}, usually extract or learn multimodal features in a concatenation fashion. Such a manner could lead to inadequate information fusion and even hurt or destroy the original components of each modality, due to the highly coupled information between multimodal RS data (particularly heterogeneous data).
\end{itemize}

According to the aforementioned challenges, we aim in this paper to build diversified multimodal RS benchmark datasets and devise novel feature learning models for land cover classification. For this purpose, we first collect multiple RS data with different resolutions, different modalities, and different sensors, with well-labeled land cover maps. Supported by these datasets, we propose a novel multimodal feature learning (MFL) model by decomposing different modalities into shared and specific representations, in order to fuse diverse information from multimodal RS data in a compacted and discriminative way. The proposed model establishes the explicit mapping relations between the to-be-learned features and original multimodal RS data, providing an interpretable MFL model. More specifically, the contributions of this paper can be highlighted as follows.
\begin{itemize}
    \item Three multimodal RS benchmark datasets are prepared and built with the application to land cover classification. They are diversified, including homogeneous HS-MS \textit{Houston2013} datasets, heterogeneous HS-SAR \textit{Berlin} datasets, and heterogeneous HS-SAR-DSM \textit{Augsburg} datasets. We will make them freely and openly available after a possible publication, contributing to the RS and information fusion communities. Currently, the heterogeneous RS feature learning and fusion techniques are less investigated, particularly three-modality datasets. To our best knowledge, this is the first time to open heterogeneous RS benchmark datasets that simultaneously involve HS, SAR, and DSM data for land cover classification.
    \item A shared and specific feature learning (S2FL) model is devised to extract diagnostic features from multimodal RS data. S2FL is capable of decoupling different modalities into shared and specific feature spaces, respectively, by aligning the common components between multi-modalities on manifolds. Such separable properties tend to capture fine-grained differences between different categories, further yielding better classification results.
    \item An alternating direction method of multipliers (ADMM)-based optimization framework is customized for the fast and accurate solutions of the proposed S2FL model. 
\end{itemize}

The rest of this paper is organized as follows. In Section II, a deep literature review is made in terms of related work in MFL. Section III then elaborates on the methodology of our proposed S2FL model as well as the ADMM-based model optimization process. In Section IV, extensive experiments are conducted on three multimodal RS benchmarks in comparison with several state-of-the-art baselines. Finally, Section V makes the summary with some important conclusions and hints at potential future research trends.

\section{Related Work}
Image-level fusion is a straightforward way to enhance certain information (e.g., spatial or spectral resolutions) of homogeneous data, such as RS image pansharpening \citep{ehlers2010multi}, HS and MS fusion \citep{wei2015hyperspectral}. However, there are more heterogeneous RS data in reality, typically optical and SAR data that can provide richer and more complementary information. The image-level fusion can not meet the demand for the heterogeneous data fusion task to some extent. Feature-level learning and fusion are needed. For decades, extensive efforts have been made by researchers to develop a variety of MFL algorithms for land cover classification of RS data. These existing methods can be roughly categorized into two main groups from the perspectives of different fusion strategies, i.e., concatenation-based MFL and alignment-based MFL.

\subsection{Concatenation-based MFL Models}
As the name suggests, concatenation-based MFL models can obtain the fused features by 
\begin{itemize}
    \item first stacking the input multimodal RS images and then passing through a certain feature extractor or learner;
    \item or first extracting or learning the feature representations for each modality and then stacking them as a certain classifier input.
\end{itemize}
There have been many classic and state-of-the-art models related to concatenation-based MFL in RS. Morphological operators \citep{fauvel2008spectral}, as the main member of the feature extractor family, have been widely and successfully applied to multimodal RS image feature extraction and classification. For instance, Liao \textit{et al.} generalized the graph embedding model \citep{hong2020graph} for the fusion of the morphological profiles of HS and LiDAR data in land cover classification \citep{liao2014generalized}. In \citep{rasti2017hyperspectral}, a novel component analysis model based on total variation was designed to further refine the feature representations of extinction profiles \citep{fang2017extinction} obtained from HS and LiDAR data. Yokoya \textit{et al.} extracted morphological features of time-series MS data and corresponding OpenStreetMaps and concatenated them as the classifier input \citep{yokoya2017multimodal}. Ma \textit{et al.} used the multisource RS data for ``Ghost City'' phenomenon identification \citep{ma2018multisource}. Authors of \citep{chen2017deep} proposed to fuse the multisource RS data by the means of deep networks for accurate land cover classification. Ref. \citep{xia2019hyperspectral} developed a semi-supervised graph fusion model for HS and LiDAR data classification. Inspired by the recent advancement of deep learning techniques in data representations, enormous learning-based approaches have been developed for multimodal RS feature fusion\citep{zhu2017}. \citep{hong2021more} for the first time proposed a general and unified deep learning framework for multimodal RS image classification. In \citep{hang2020classification}, a coupled convolutional neural network was employed to fuse the heterogeneous RS data in the feature level. 
However, there is room for improvement in the interpretablity of DL-based methods. The interpretable knowledge embedding can guide the network learning towards better solutions (or results) more effectively. Furthermore, the past decades have witnessed a favorable development of concatenation-based MFL in the RS community, yet the ability to fully take advantage of the diverse information of different modalities, especially heterogeneous data, remains limited. 

\subsection{Alignment-based MFL Models}
Unlike the above feature concatenation strategy, alignment-based MFL methods seek to learn a common feature set that multiple modalities share by the means of well-known manifold alignment (MA) techniques \citep{wang2011heterogeneous}. By introducing the MA into the RS applications, Tuia \textit{et al.} aligned the multi-view MS data to a consistent representation in a semi-supervised way \citep{tuia2014semisupervised}. \citep{tuia2016kernel} projected the multimodal data to a higher-dimensional kernel space, where different data sources can be better aligned. Moreover, the semi-supervised model in \citep{tuia2014semisupervised} was improved by using a mapper-induced graph structure for the fusion of optical image and polarimetric SAR data \citep{hu2019mima}. Inspired by the MA idea, Hong \textit{et al.} proposed a MA-regularized representation learning model, simultaneously involving subspace learning and ridge regression, CoSpace for short \citep{hong2019cospace}. CoSpace is capable of learning the alignment representations across multi-modalities, thereby yielding more effective information fusion. The same investigators further extended the CoSpace model by replacing ridge regression with sparse regression, generating a $\ell_1$-norm version of the CoSpace model, i.e., $\ell_1$-CoSpace \citep{hong2020learning}. Pournemat \textit{et al.} \citep{pournemat111semisupervised} proposed a semi-supervised charting approach for multimodal MA in the spectral domain. There are also some MA-based variants successfully applied to other RS-related applications, such as visualization \citep{liao2016manifold}, dimensionality reduction, cross-modality retrieval \citep{hong2019learnable}. Admittedly, the alignment-based strategy is capable of performing well in heterogeneous data fusion by the means of information-sharing mechanism. Yet only capturing the shared information across multi-modalities is hardly achievable to learn better multimodal feature representations, due to the lack of modeling or fusing modality-specific properties.

\begin{figure*}[!t]
	  \centering
			\includegraphics[width=0.9\textwidth]{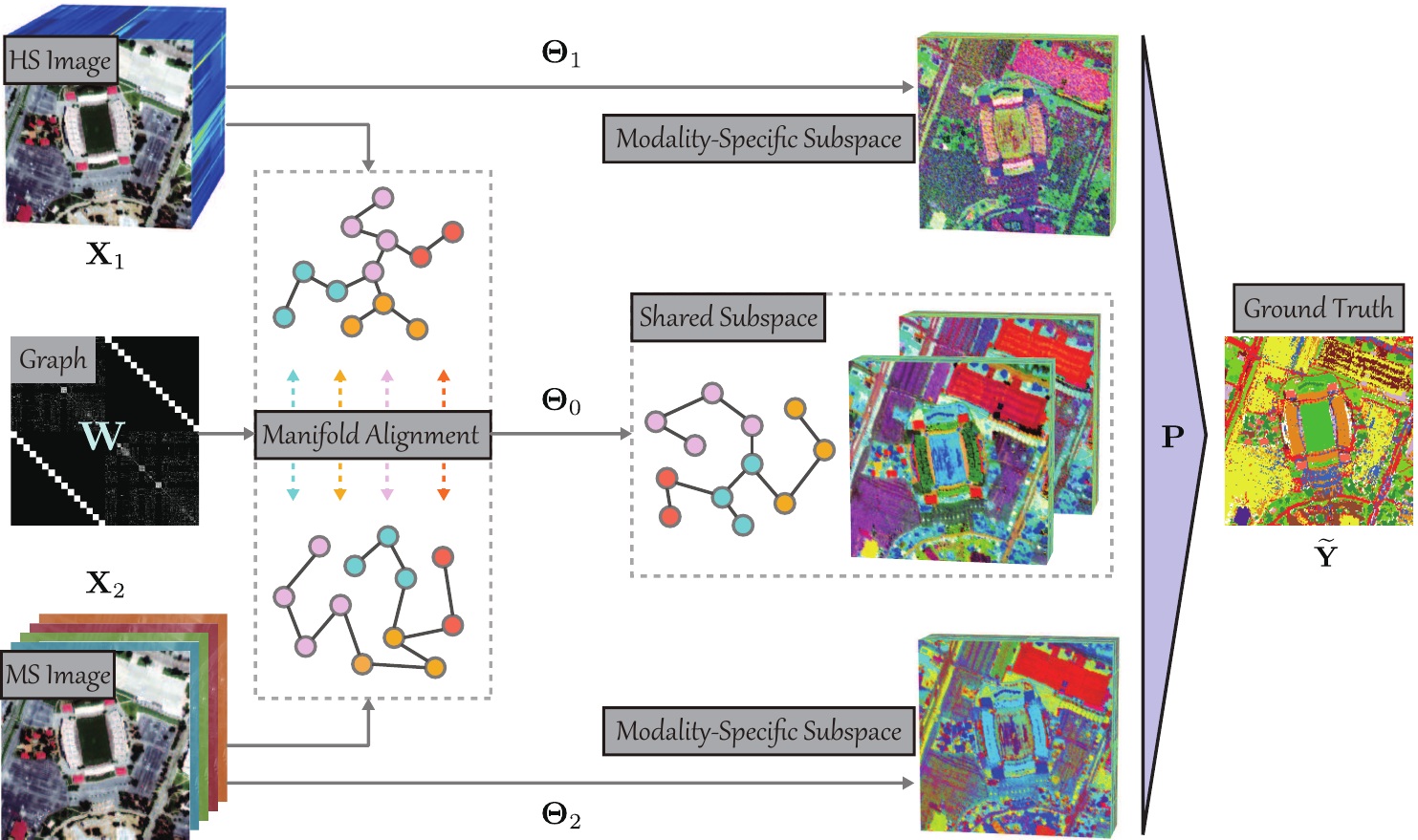}
        \caption{An illustration to clarify the learning process for shared and specific subspaces (or features) of multimodal RS data in the proposed S2FL model. The to-be-estimated variables $\mathbf{\Theta}_{0}$, $\{\mathbf{\Theta}_{k}\}_{k=1}^{2}$, and $\mathbf{P}$ denote the shared subspace projection, the specific subspace projections, and the regression matrix, respectively, Note that we here take the bi-modality as an example.}
\label{fig:workflow_IF}
\end{figure*}

\section{S2FL: Shared and Specific Feature Learning Model}

\subsection{Method Overview}
To enhance the representation ability of multimodal data fusion and reduce the information loss (possibly due to only considering modality-shared properties and ignoring those modality-specific ones), we seek to find a more discriminative feature space by disentangling different data sources into shared and specific domains. Using the to-be-learned features, a better decision boundary is expect to obtain in the classification task. For this purpose, a shared and specific feature learning model is devised, called S2FL, by aligning shared components between multi-modalities on the latent manifold subspace and simultaneously separating out their specific information. Such a modeling strategy is interpretable and effective for learning multimodal RS feature representations, further yielding the great potentials in land cover classification. Fig. \ref{fig:workflow_IF} illustrates the flowchart of the proposed S2FL model.

\subsection{Notation}
Let $\mathbf{X}_{k}\in \mathbb{R}^{d_{k}\times N}$ be the unfolded matrix with respect to the $k^{th}$ modality with $d_{k}$ channels by $N$ pixels, and $K$ be the number of all considered modalities. $\mathbf{Y}\in \mathbb{R}^{C\times N}$ denotes the one-hot encoding label matrix, where $C$ is the number of categories. $\mathbf{\Theta}_{0}\in \mathbb{R}^{d_{s}\times \sum_{k=1}^{K}d_{k}}$ and $\mathbf{\Theta}_{k}\in \mathbb{R}^{d_{s}\times d_{k}}$ denote the shared subspace projection and specific subspace projections with the respect to the $k^{th}$ modality, respectively, and $d_{s}$ means the feature (or subspace) dimension. $\mathbf{P}\in \mathbb{R}^{C\times d_{s}}$ is defined as the regression matrix that connects the subspace and label information (i.e., $\mathbf{Y}$). $\mathbf{I}$, $\norm{\mathbf{X}}_{\F}$, and $\tr(\mathbf{X})$ denote the identity matrix, the Frobenius norm of the matrix $\mathbf{X}$, and the trace of the matrix $\mathbf{X}$, respectively. Moreover, the Laplacian matrix is denoted by $\mathbf{L}$, which can be computed by $\mathbf{D}-\mathbf{W}$. $\mathbf{W}$ is the adjacency matrix of $\mathbf{X}$, and $\mathbf{D}$ is a diagonal matrix, whose diagonal elements can be obtained by $\mathbf{D}_{ii}=\sum_{i\neq j}\mathbf{W}_{i,j}$. Table \ref{table:notations} gives the detailed definitions of these variables used in the S2FL model.

\begin{table*}[!t]
\centering
\caption{The symbols of variables used in the proposed S2FL model as well as their description and size, where we take the bi-modality as an example, i.e., $K=2$.}
\vspace{2mm}
\resizebox{1\textwidth}{!}{ 
\begin{tabular}{c||c|c}
\toprule[1.5pt]
Symbols & Description & Size\\
\hline \hline
$d_{s}$ & the dimension of the feature space or subspace & $1\times 1$ \\
$d_{k}$ & the dimension of the $k^{th}$ modality & $1\times 1$ \\
$C$ & the number of categories & $1\times 1$\\
$N$ & the number of pixels (or samples) & $1\times 1$\\
$K$ & the number of considered modalities & $1\times 1$\\
$\mathbf{\cdot}_{i,j}$ & the $(i,j)^{th}$ entry of the matrix $\mathbf{\cdot}$ & $1\times 1$ \\
$\mathbf{X}_{k}$ & the unfolded matrix of the $k^{th}$ modality & $d_{k}\times N$\\
$\mathbf{Y}$ & the one-hot encoding label matrix & $C\times N$\\
$\mathbf{\Theta}_{0}$ & the shared subspace projections for all considered modalities & $d_{s}\times \sum_{k=1}^{K}d_{k}$\\
$\mathbf{\Theta}_{k}$ & the specific subspace projection for the $k^{th}$ modality & $d_{s}\times d_{k}$\\
$\mathbf{\Theta}$ & the generalized subspace projections, obtained by $\mathbf{\Theta}_{0} + [\mathbf{\Theta}_{1},\mathbf{\Theta}_{2}]$ & $d_{s}\times \sum_{k=1}^{K}d_{k}$\\
$\mathbf{P}$ & the linear regression matrix & $C\times d_{s}$\\
$\mathbf{W}$ & the adjacency matrix of to-be-aligned modalities & $2N\times 2N$\\
$\mathbf{D}$ & the degree matrix of the matrix $\mathbf{W}$, obtained by $\mathbf{D}_{ii}=\sum_{i\neq j}\mathbf{W}_{i,j}$ & $2N\times 2N$\\
$\mathbf{L}$ & the Laplacian matrix of the matrix $\mathbf{W}$, obtained by $\mathbf{D}-\mathbf{W}$ & $2N\times 2N$\\
$\mathbf{I}$ & the identity matrix & $d_{s}\times d_{k}$\\
$\norm{\mathbf{X}}_{\F}$ & the Frobenius norm of the matrix $\mathbf{X}$, obtained by $\sqrt{\sum_{i,j}\mathbf{X}_{i,j}^{2}}$ & $1\times 1$ \\
$\tr(\mathbf{X})$ & the trace of the matrix $\mathbf{X}$ & $1\times 1$ \\
\bottomrule[1.5pt]
\end{tabular}
}
\label{table:notations}
\end{table*}

\subsection{Problem Formulation}
With the aforementioned problem statement and given definitions of variables, we model the S2FL's problem as follows
\begin{equation}
\label{eq1}
\begin{aligned}
       \mathop{\min}_{\mathbf{P},\mathbf{\Theta}_{0}, \{\mathbf{\Theta}_{k}\}_{k=1}^{K}} & \frac{1}{2}\norm{\widetilde{\mathbf{Y}}-\mathbf{P}\mathbf{\Theta}\widetilde{\mathbf{X}}}_{\F}^{2}+\frac{\alpha}{2}\norm{\mathbf{P}}_{\F}^{2}+\frac{\beta}{2}\tr(\mathbf{\Theta}_{0}\widetilde{\mathbf{X}}\mathbf{L}(\mathbf{\Theta}_{0}\widetilde{\mathbf{X}})^{\top})\\
       \mathrm{s.t.} & \;\; \mathbf{\Theta}_{k}\mathbf{\Theta}_{k}^{\top}=\mathbf{I},\;\; k=0,1,2,...K,
\end{aligned}
\end{equation}
where 
\begin{equation*}
\begin{aligned}
      \widetilde{\mathbf{Y}}=
      \begin{bmatrix}
             \mathbf{Y}_{1}, & \cdots, & \mathbf{Y}_{K}
      \end{bmatrix},\;\;
      \widetilde{\mathbf{X}}=
      \begin{bmatrix}
             \mathbf{X}_{1}, & \cdots, & \mathbf{0} \\
             \vdots & \ddots & \vdots \\
             \mathbf{0}, & \cdots, & \mathbf{X}_{K}
      \end{bmatrix}, \;\;
      \mathbf{\Theta}= \mathbf{\Theta}_{0} + 
      \begin{bmatrix}
             \mathbf{\Theta}_{1}, & \cdots, & \mathbf{\Theta}_{K}
      \end{bmatrix}.
\end{aligned}
\end{equation*}
$\alpha$ and $\beta$ denote the penalty parameters to balance the importance of different terms. Please note that $\mathbf{Y}_{1}$, $\mathbf{Y}_{2}$, $...$, and $\mathbf{Y}_{K}$ only aim to show the groups from $1$ to $K$ corresponding to different modalities, and actually represent the same labels, i.e., $\mathbf{Y}$.

The optimization problem of the first term in Eq. (\ref{eq1}) is highly ill-posed due to its large freedom degrees (e.g., simultaneous estimation of coupled variables $\mathbf{P}$ and $\mathbf{\Theta}$). To this end, 

\begin{figure*}[!t]
	  \centering
			\includegraphics[width=0.55\textwidth]{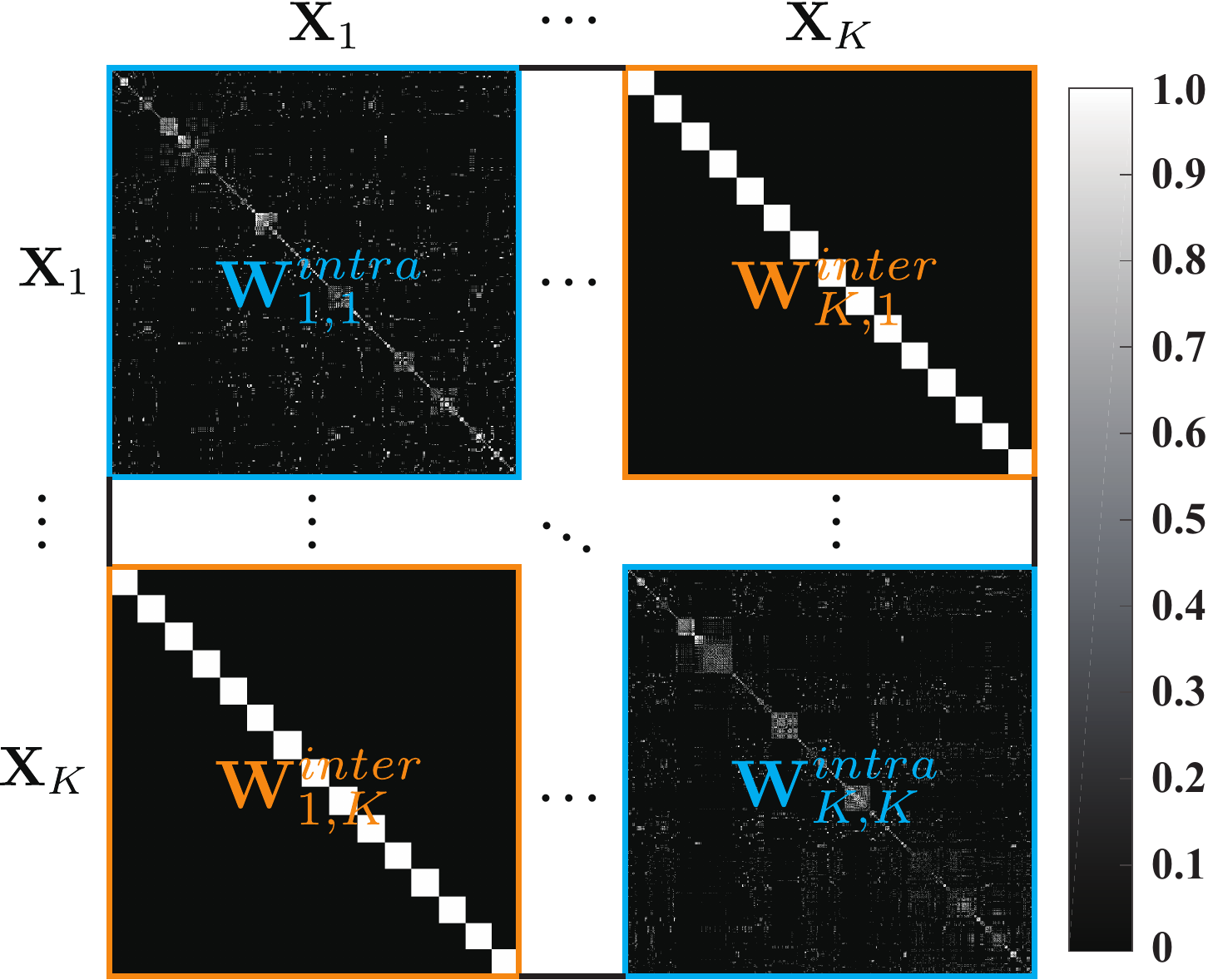}
        \caption{An example to illustrate the adjacency matrix ($\mathbf{W}$).}
\label{fig:graph}
\end{figure*}

\begin{itemize}
    \item we regularize the variable $\mathbf{P}$ with its Frobenius norm, denoted as $\norm{\mathbf{P}}_{\F}$, in order to stabilize the convergence process and improve the generalization ability of the model;
    \item the modality-shared and modality-specific components can be learned and separated on a latent subspace by the means of one common projection $\mathbf{\Theta}_{0}$ for all modalities and $K$ characteristic projections $\{\mathbf{\Theta}_{k}\}_{k=1}^{K}$ for each individual modality. This information-sharing process can be performed by using the MA regularization, i.e., $\tr(\mathbf{\Theta}_{0}\widetilde{\mathbf{X}}\mathbf{L}(\mathbf{\Theta}_{0}\widetilde{\mathbf{X}})^{\top})$, and then the remaining information is naturally irrelevant and unique between different modalities. Moreover, the joint graph $\mathbf{W}$ in $\mathbf{L}$ consists of $k^{2}$ subgraphs, as illustrated in Fig. \ref{fig:graph}. In $\mathbf{W}$, the block diagonal matrix is the intra-modality subgraph, which can be obtained by using the Gaussian kernel function with the width of $\sigma$ as follows
    \begin{equation}
        \label{eq2}
        \begin{aligned}
            \mathbf{W}_{i,j}^{intra} = \mathrm{exp}(-\frac{\norm{\mathbf{x}_{i}-\mathbf{x}_{j}}^{2}}{\sigma^{2}}),
        \end{aligned}
    \end{equation}
    and the rest is the inter-modality subgraph, which can be directly constructed by the means of label information, leading to the following discriminative graph structure \citep{hong2019cospace}:
    \begin{equation}
    \label{eq3}
      \mathbf{W}_{i,j}^{inter}=
        \begin{cases}
          \begin{aligned}
          1/N_{C}, \quad & \text{if \(\mathbf{x}_{i}\) and \(\mathbf{x}_{j}\) belong to the $C^{th}$ class;}\\
          0, \quad & \text{otherwise,}
          \end{aligned}
        \end{cases}
    \end{equation}
    where $N_{C}$ is the number of samples for the $C^{th}$ class;
    \item the orthogonal constraints with respect to the variables $\mathbf{\Theta}_{0}$ and $\{\mathbf{\Theta}_{k}\}_{k=1}^{K}$ are added in S2FL model to reduce the freedoms and shrink the solution space effectively, thereby finding better local optimal solutions. 
\end{itemize}

\begin{algorithm}[!t]
\caption{S2FL: Global optimization}
\begin{algorithmic}[1]
\Require $\widetilde{\mathbf{Y}}$, $\widetilde{\mathbf{X}}$, $\mathbf{L}$, and parameters $\alpha$, $\beta$, $\sigma$, and $\mathrm{maxIter}$.
\State Initialize the parameter $\mathbf{\Theta}$ using locality preserving projections (LPP) \citep{he2004locality} and the parameter $\mathbf{\Theta}_{0}=\mathbf{0}$. The iteration starts with $t=1$ and the tolerate error is set to $\zeta=10^{-4}$.
\State Compute the adjacency matrix $\mathbf{W}$ using Eqs. (\ref{eq2}) and (\ref{eq3}) Laplacian matrix $\mathbf{L}$.
\For {The iteration number $t=1$ to $\mathrm{maxIter}$}
    \State Learn the linear regression matrix $\mathbf{P}$ using Eq. (\ref{eq5}).
    \State Learn the shared subspace projection matrix $\mathbf{\Theta}_{0}$ by \textbf{Algorithm 2}.
    \For {$k=1$ to $K$}
         \State Learn the specific subspace projection $\mathbf{\Theta}_{k}$ for the $k^{th}$ modality.
    \EndFor
    \State When $t>1$ and calculate the loss of the objective function in the $t^{th}$ and $(t-1)^{th}$ iterations, denoted as $E^{t}$ and $E^{t-1}$, and check the stopping condition:
    \If{$|\frac{E^{t}-E^{t-1}}{E^{t-1}}|<\zeta$}
    \State Stop iteration.
    \Else
    \State $t\leftarrow t+1$.
    \EndIf
    \EndFor
\Ensure Linear regression matrix $\mathbf{P}$, shared subspace projection $\mathbf{\Theta}_{0}$ for all considered modalities, and modality-specific subspace projections $\{\mathbf{\Theta}_{k}\}_{k=1}^{K}$.
\end{algorithmic}
\end{algorithm}

\subsection{Model Optimization}
The optimization problem of Eq. (\ref{eq1}) is typically non-convex, whose global minimum is usually hard to be found. We, however, expect to have local optimal solutions by alternatively optimizing separable convex subproblems of to-be-estimated variables $\mathbf{P}$, $\mathbf{\Theta}_{0}$, and $\{\mathbf{\Theta}_{k}\}_{k=1}^{K}$. \textbf{Algorithm 1} details an overview optimization process of the proposed S2FL model.

\subsubsection{Learning Linear Regression Matrix --- $\mathbf{P}$}
The optimization with respect to the variable $\mathbf{P}$ is nothing but a least-square regression problem with a common Tikhonov-Phillips regularization. This subproblem can be then written as
\begin{equation}
\label{eq4}
\begin{aligned}
       \mathop{\min}_{\mathbf{P}}\frac{1}{2}\norm{ \widetilde{\mathbf{Y}}-\mathbf{P}\mathbf{\Theta}\widetilde{\mathbf{X}}}_{\F}^{2}+\frac{\alpha}{2}\norm{\mathbf{P}}_{\F}^{2},
\end{aligned}
\end{equation}
which has the following analytical solution
\begin{equation}
\label{eq5}
\begin{aligned}
        \mathbf{P} \leftarrow ( \widetilde{\mathbf{Y}}\widetilde{\mathbf{X}}^{\top}\mathbf{\Theta}^{\top})(\mathbf{\Theta}\widetilde{\mathbf{X}}\widetilde{\mathbf{X}}^{\top}\mathbf{\Theta}^{\top}+\alpha\mathbf{I})^{-1}.
\end{aligned}
\end{equation}

\subsubsection{Learning Shared Subspace Projection Matrix --- $\mathbf{\Theta}_{0}$}
The constraint optimization problem with respect to the variable $\mathbf{\Theta}_{0}$ can be formulated as follows 
\begin{equation}
\label{eq6}
\begin{aligned}
 \mathop{\min}_{\mathbf{\Theta}_{0}} \frac{1}{2}\norm{ \widetilde{\mathbf{Y}}-\mathbf{P}\mathbf{\Theta}\widetilde{\mathbf{X}}}_{\F}^{2}+\frac{\beta}{2}\tr(\mathbf{\Theta}_{0}\mathbf{\widetilde{X}}\mathbf{L}(\mathbf{\Theta}_{0}\mathbf{\widetilde{X}})^{\top})\;\; \mathrm{s.t.} \;\; \mathbf{\Theta}_{0}\mathbf{\Theta}_{0}^{\top}=\mathbf{I}.
\end{aligned}
\end{equation}
The problem (\ref{eq6}) can be optimized by designing an ADMM-based solver \citep{boyd2011distributed}. More specifically, two auxiliary variables $\mathbf{H}$ and $\mathbf{G}$ are introduced into the Eq. (\ref{eq6}) to replace $\mathbf{\Theta}_{0}\widetilde{\mathbf{X}}$ in the first term and $\mathbf{\Theta}_{0}$ in the constraint term, respectively, we then have the following equivalent form of Eq. (\ref{eq6})
\begin{equation}
\label{eq7}
\begin{aligned}
\mathop{\min}_{\mathbf{\Theta}_{0},\mathbf{H},\mathbf{G}}&\frac{1}{2}\norm{ \widetilde{\mathbf{Y}}-\mathbf{P}[\mathbf{\Theta}_{1},\cdots, \mathbf{\Theta}_{K}]\widetilde{\mathbf{X}}-\mathbf{P}\mathbf{H}}_{\F}^{2}+\frac{\beta}{2}\tr(\mathbf{\Theta}_{0}\widetilde{\mathbf{X}}\mathbf{L}(\mathbf{\Theta}_{0}\widetilde{\mathbf{X}})^{\top})\\
&\mathrm{s.t.} \;\; \mathbf{H}=\mathbf{\Theta}_{0}\widetilde{\mathbf{X}}, \;\; \mathbf{G}=\mathbf{\Theta}_{0}, \;\; \mathbf{G}\mathbf{G}^{\top}=\mathbf{I}.
\end{aligned}
\end{equation}
We further rewrite the Eq. (\ref{eq6}) to its augmented Lagrangian function:
\begin{equation}
\label{eq8}
\begin{aligned}
       \mathscr{L}&\left(\mathbf{\Theta}_{0}, \mathbf{H}, \mathbf{G}, \mathbf{\Lambda}_{1}, \mathbf{\Lambda}_{2} \right)\\
       &=\frac{1}{2}\norm{ \widetilde{\mathbf{Y}}-\mathbf{P}[\mathbf{\Theta}_{1},\cdots, \mathbf{\Theta}_{K}]\widetilde{\mathbf{X}}-\mathbf{P}\mathbf{H}}_{\F}^{2}+\frac{\beta}{2}\tr(\mathbf{\Theta}_{0}\widetilde{\mathbf{X}}\mathbf{L}(\mathbf{\Theta}_{0}\widetilde{\mathbf{X}})^{\top})\\
       &+\mathbf{\Lambda}_{1}^{\top}(\mathbf{H}-\mathbf{\Theta}_{0}\widetilde{\mathbf{X}})+\mathbf{\Lambda}_{2}^{\top}(\mathbf{G}-\mathbf{\Theta}_{0})+\frac{\mu}{2}\norm{\mathbf{H}-\mathbf{\Theta}_{0}\widetilde{\mathbf{X}}}_{\F}^{2}+\frac{\mu}{2}\norm{\mathbf{G}-\mathbf{\Theta}_{0}}_{\F}^{2}\\
       &\;\;\mathrm{s.t.} \;\; \mathbf{G}\mathbf{G}^{\top}=\mathbf{I},
\end{aligned}
\end{equation}
where the variables $\mathbf{{\Lambda}_{1}}$ and $\mathbf{{\Lambda}_{2}}$ denote the Lagrange multipliers and $\mu$ is the regularization parameter.

Under the ADMM optimization framework, the problem (\ref{eq8}) can be effectively solved by successively minimizing the object function $\mathscr{L}$ for the variables $\mathbf{\Theta}_{0}$, $\mathbf{H}$, $\mathbf{G}$, $\mathbf{\Lambda}_{1}$, and $\mathbf{\Lambda}_{2}$, respectively, when other variables are fixed. 

\textit{Optimization with respect to $\mathbf{\Theta}_{0}$:} The optimization problem of the variable $\mathbf{\Theta}_{0}$ can be formulated as follows
\begin{equation}
\label{eq9}
\begin{aligned}
       \mathop{\min}_{\mathbf{\Theta}_{0}}&\frac{\beta}{2}\tr(\mathbf{\Theta}_{0}\widetilde{\mathbf{X}}\mathbf{L}(\mathbf{\Theta}_{0}\widetilde{\mathbf{X}})^{\top})+\mathbf{\Lambda}_{1}^{\top}(\mathbf{H}-\mathbf{\Theta}_{0}\widetilde{\mathbf{X}})+\mathbf{\Lambda}_{2}^{\top}(\mathbf{G}-\mathbf{\Theta}_{0})\\
       &+\frac{\mu}{2}\norm{\mathbf{H}-\mathbf{\Theta}_{0}\widetilde{\mathbf{X}}}_{\F}^{2}+\frac{\mu}{2}\norm{\mathbf{G}-\mathbf{\Theta}_{0}}_{\F}^{2},
\end{aligned}
\end{equation}
hence its closed-form solution is 
\begin{equation}
\label{eq10}
\begin{aligned}
       \mathbf{\Theta}_{0}\leftarrow(\mu\mathbf{H}\widetilde{\mathbf{X}}^{\top}+\mathbf{\Lambda}_{1}\widetilde{\mathbf{X}}^{\top}+\mu\mathbf{G}+\mathbf{\Lambda}_{2})\times(\mu\widetilde{\mathbf{X}}\widetilde{\mathbf{X}}^{\top}+\mu\mathbf{I}+\beta\widetilde{\mathbf{X}}\mathbf{L}\widetilde{\mathbf{X}}^{\top})^{-1}.
\end{aligned}
\end{equation}

\textit{Optimization with respect to $\mathbf{H}$:} We estimate the variable $\mathbf{H}$ by solving the following optimization problem:
\begin{equation}
\label{eq11}
\begin{aligned}
      \mathop{\min}_{\mathbf{H}}\frac{1}{2}\norm{ \widetilde{\mathbf{Y}}-\mathbf{P}[\mathbf{\Theta}_{1},\cdots, \mathbf{\Theta}_{K}]\widetilde{\mathbf{X}}-\mathbf{P}\mathbf{H}}_{\F}^{2}+\mathbf{\Lambda}_{1}^{\top}(\mathbf{H}-\mathbf{\Theta}_{0}\widetilde{\mathbf{X}})+\frac{\mu}{2}\norm{\mathbf{H}-\mathbf{\Theta}_{0}\widetilde{\mathbf{X}}}_{\F}^{2}.
\end{aligned}
\end{equation}
For Eq. (\ref{eq11}), it is straightforward to derive the analytical solution, i.e., 
\begin{equation}
\label{eq12}
\begin{aligned}
       \mathbf{H}\leftarrow(\mathbf{P}^{\top}\mathbf{P}+\mu\mathbf{I})^{-1}(\mathbf{P}^{\top}(\widetilde{\mathbf{Y}}-\mathbf{P}[\mathbf{\Theta}_{1},\cdots, \mathbf{\Theta}_{K}]\widetilde{\mathbf{X}})+\mu\mathbf{\Theta}_{0}\widetilde{\mathbf{X}}-\mathbf{\Lambda}_{1}).
\end{aligned}
\end{equation}

\begin{algorithm}[!t]
\caption{Subprobelm optimization with respect to the variable $\mathbf{\Theta}_{0}$}
\begin{algorithmic}[1]
\Require $\widetilde{\mathbf{Y}}$, $\mathbf{P}$, $\widetilde{\mathbf{X}}$, $\mathbf{L}$, and the parameter $\beta$, and $\mathrm{maxIter}$.
\State Initialize $\mathbf{\Theta}_{0}=\mathbf{0}$, the auxiliary variables, e.g., $\mathbf{G}=\mathbf{\Lambda}_{2}=\mathbf{0}$, $\mathbf{\Lambda}_{1}=\mathbf{0}$. The regularization parameter is $\mu=10^{-3}$, whose maximal limitation and scaling factor are $\mu_{\max}=10^{6}$ and $\rho=1.5$, respectively. The iteration starts with $t=1$ and the tolerate error of the variable is set to $\varepsilon=10^{-6}$.
\For {The iteration number $t=1$ to $\mathrm{maxIter}$}
    \State Fix other variables and update $\mathbf{H}$ using Eq. (\ref{eq12}).
    \State Fix other variables and update $\mathbf{\Theta}_{0}$ using Eq. (\ref{eq10}).
    \State Fix other variables and update $\mathbf{G}$ by the SOC solver, i.e., Eqs. (\ref{eq14}) and (\ref{eq15}).
    \State Update Lagrange multipliers $\mathbf{\Lambda}_{1}$ and $\mathbf{\Lambda}_{2}$ using Eq. (\ref{eq16}).
    \State Update the regularization parameter using Eq. (\ref{eq17}).
    \State Check the stopping condition:
    \If{$\norm {\mathbf{H}-\mathbf{\Theta}_{0}\widetilde{\mathbf{X}}}_{\F}<\varepsilon$ and $\norm {\mathbf{G}-\mathbf{\Theta}_{0}}_{\F}<\varepsilon$}
    \State Stop iteration.
    \Else
    \State $t\leftarrow t+1$.
    \EndIf
    \EndFor
\Ensure Shared subspace projection $\mathbf{\Theta}_{0}$.
\end{algorithmic}
\end{algorithm}

\textit{Optimization with respect to $\mathbf{G}$:} The optimization problem with the orthogonal constraint for the variable $\mathbf{G}$ is 
\begin{equation}
\label{eq13}
\begin{aligned}
\mathop{\min}_{\mathbf{G}}\mathbf{\Lambda}_{2}^{\top}(\mathbf{G}-\mathbf{\Theta}_{0})+\frac{\mu}{2}\norm{\mathbf{G}-\mathbf{\Theta}_{0}}_{\F}^{2} \;\; \mathrm{s.t.} \;\; \mathbf{G}\mathbf{G}^{\top}=\mathbf{I},
\end{aligned}
\end{equation}
whose solution can be obtained by the means of a well-known solver, i.e., splitting orthogonality constraints (SOC) \citep{lai2014splitting}. The method of SOC solves the orthogonality constrained problem in two steps.
\begin{itemize}
    \item 1) Singular value decomposition (SVD) is performed on the variable $\mathbf{\Theta}_{0}$, i.e.,
    \begin{equation}
        \label{eq14}
        \begin{aligned}
              \left[\mathbf{U},\mathbf{\Sigma},\mathbf{V}\right]=\svd(\mathbf{\mathbf{\Theta}_{0}-\mathbf{\Lambda}_{2}/\mu}),
        \end{aligned}
    \end{equation}
    such that $\mathbf{\mathbf{\Theta}_{0}-\mathbf{\Lambda}_{2}/\mu}=\mathbf{U}\mathbf{\Sigma}\mathbf{V}^{\top}$.
    \item 2) The orthogonality is satisfied when the variable $\mathbf{G}$ is updated by
    \begin{equation}
        \label{eq15}
        \begin{aligned}
              \mathbf{G}\leftarrow\mathbf{U}\mathbf{I}_{n \times m}\mathbf{V}^{\top}.
        \end{aligned}
    \end{equation}
\end{itemize}

\textit{Updating with respect to Lagrange multipliers $\mathbf{\Lambda}_{1}$ and $\mathbf{\Lambda}_{2}$:}
\begin{equation}
\label{eq16}
\begin{aligned}
       \mathbf{\Lambda}_{1}\leftarrow\mathbf{\Lambda}_{1}+\mu (\mathbf{H}-\mathbf{\Theta}_{0}\widetilde{\mathbf{X}}),\qquad \mathbf{\Lambda}_{2}\leftarrow\mathbf{\Lambda}_{2}+\mu (\mathbf{G}-\mathbf{\Theta}_{0}).
\end{aligned}
\end{equation}

\textit{Updating with respect to the regularization parameter $\mu$:}
\begin{equation}
\label{eq17}
\begin{aligned}
       \mu\leftarrow\ \min (\rho\mu, \mu_{\max}),
\end{aligned}
\end{equation}
where $\rho>1$ and $\mu_{\max}$ denote the scaling factor and the maximal value of $\mu$ within finite steps, respectively.

The specific optimization procedures for estimating the variable $\mathbf{\Theta}_{0}$, i.e., solving the problem (\ref{eq6}), are summarized in \textbf{Algorithm 2}.

\subsubsection{Learning Specific Subspace Projection Matrices --- $\{\mathbf{\Theta}_{k}\}_{k=1}^{K}$}
The solution to the variable $\mathbf{\Theta}_{k}$ can be obtained by using the same solver with the problem (\ref{eq6}). The only difference lies in that there is no the MA term, i.e., $\frac{\beta}{2}\tr(\mathbf{\Theta}_{0}\widetilde{\mathbf{X}}\mathbf{L}(\mathbf{\Theta}_{0}\widetilde{\mathbf{X}})^{\top})$, in the optimization problem of $\{\mathbf{\Theta}_{k}\}_{k=1}^{K}$. As a result, the \textbf{Algorithm 2} can be directly applied to estimate the variables $\{\mathbf{\Theta}_{k}\}_{k=1}^{K}$ as well.

\begin{figure*}[!t]
	  \centering
	  	\subfigure{
			\includegraphics[width=0.325\textwidth]{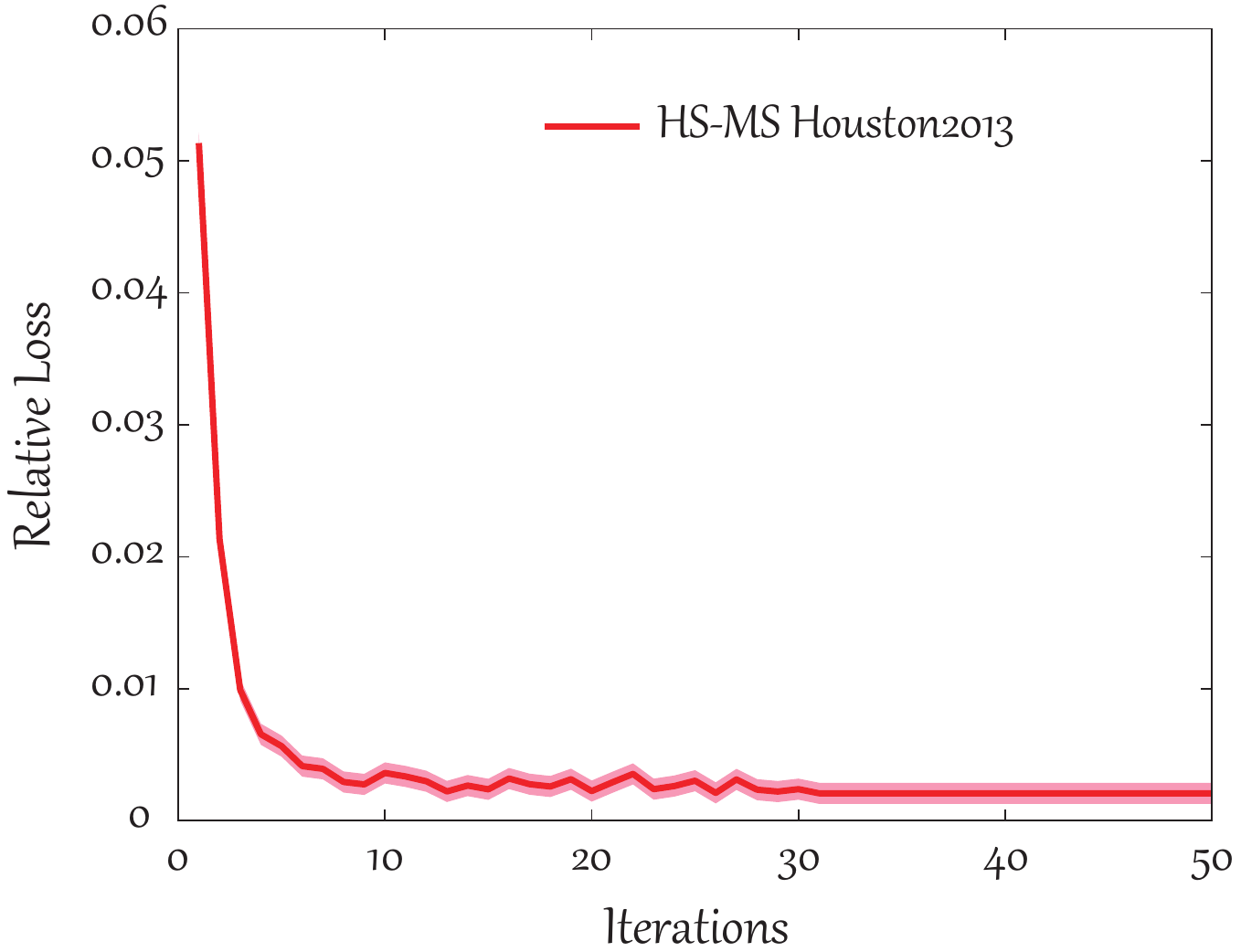}
		}\hspace{-9pt}
	  	\subfigure{
			\includegraphics[width=0.32\textwidth]{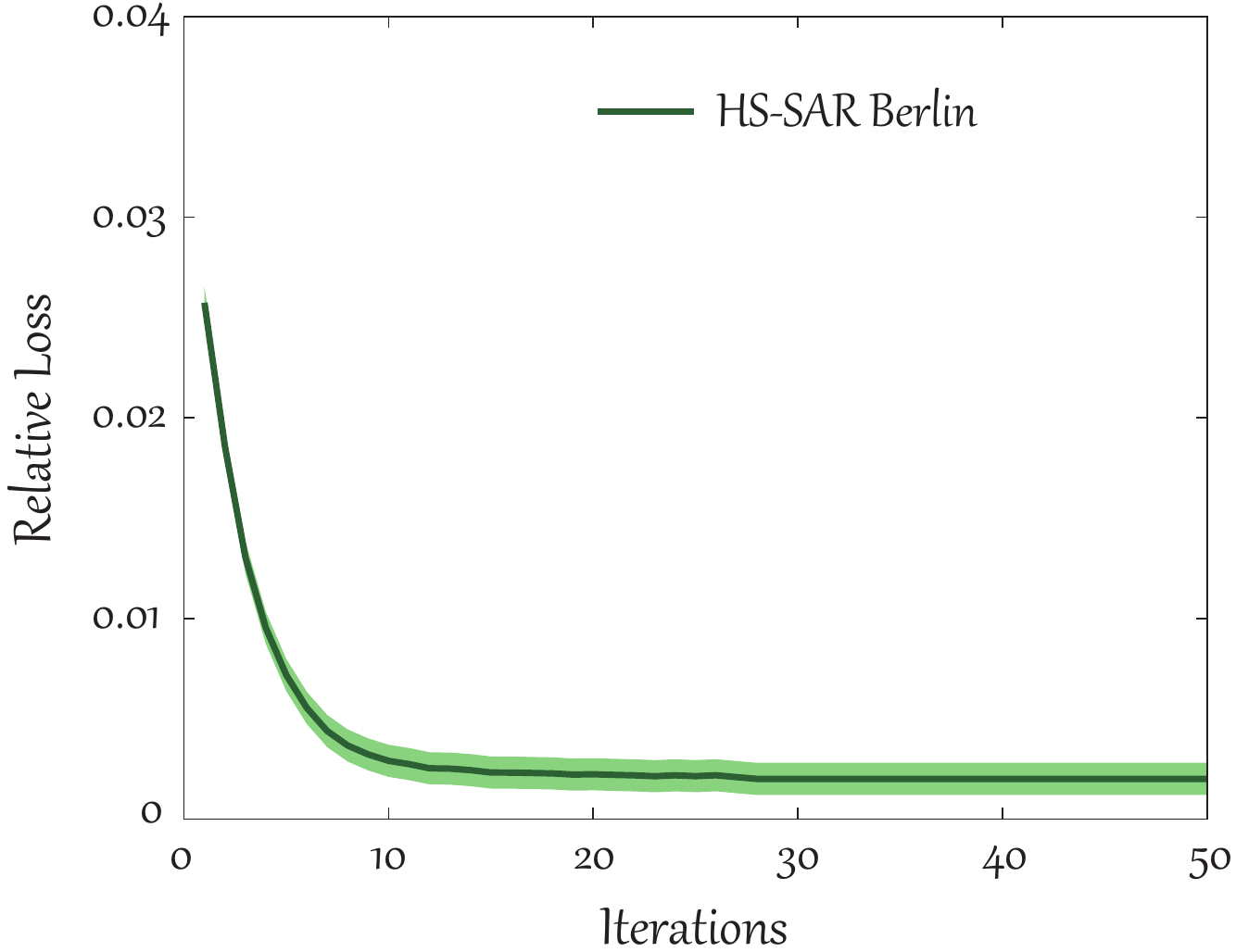}
		}\hspace{-9pt}
		\subfigure{
			\includegraphics[width=0.325\textwidth]{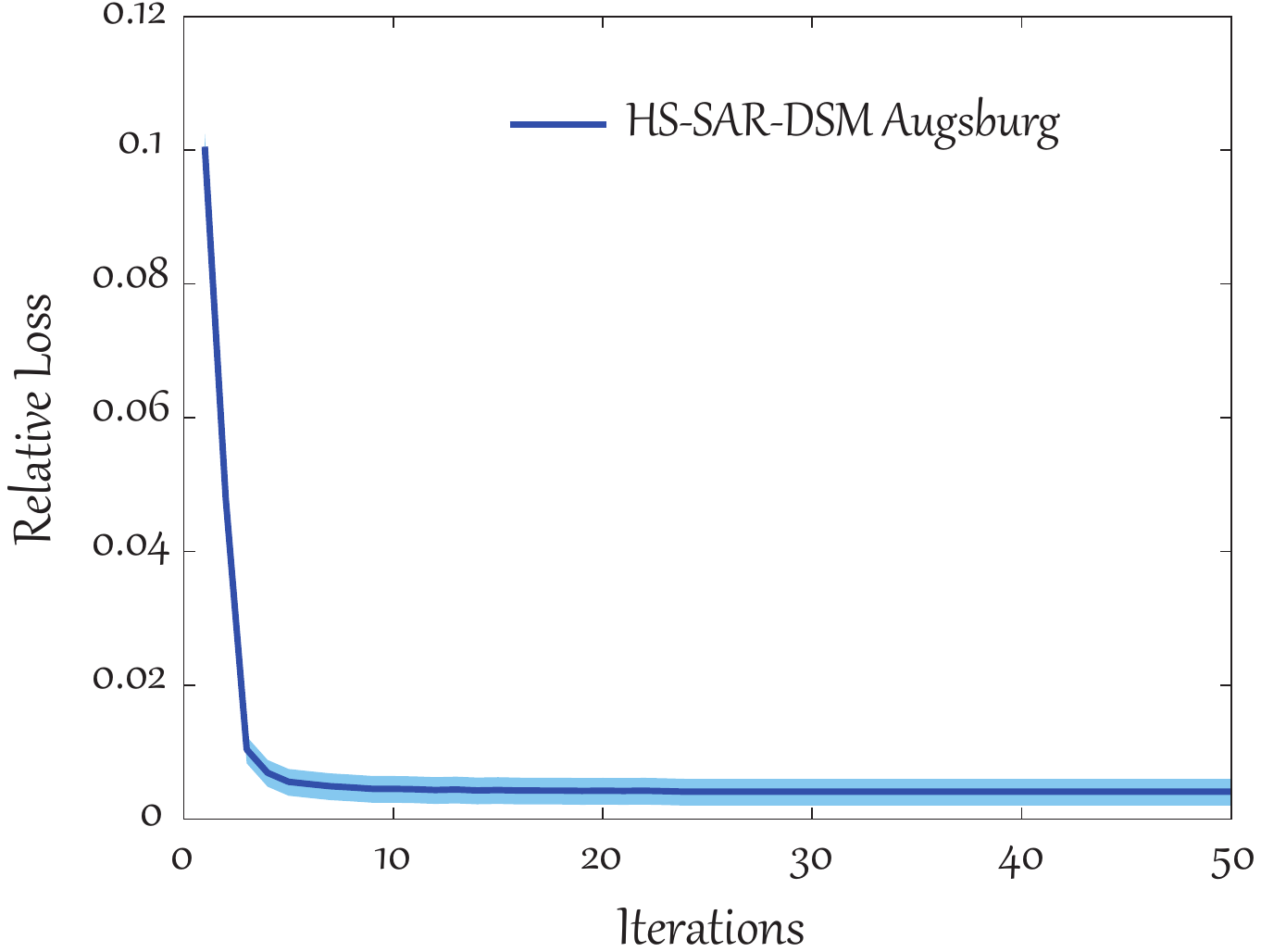}
		}
        \caption{Convergence analysis of the S2FL model on three multimodal RS benchmark datasets: HS-MS \textit{Houston2013}, HS-SAR \textit{Berlin}, and HS-SAR-DSM \textit{Augsburg}.}
\label{fig:Conv_curves}
\end{figure*}

\subsection{Convergence Analysis and Computational Cost}
The global optimization given in \textbf{Algorithm 1} can be performed by the alternating minimization. The block coordinate descent (BCD) is a commonly-used strategy to solve the problem. The BCD is able to converge well in theory as long as the convexity for each subproblem is met \citep{bazaraa2013nonlinear}. The ADMM solver in \textbf{Algorithm 2} is actually a variant of \emph{inexact} Augmented Lagrange Multiplier (ALM) \citep{lin2010augmented}, which has been successfully used for solving the multi-block ADMM-based optimization problems \citep{chen2016direct,deng2017parallel,wang2018convergence}. Up to the present, the convergence of the multi-block ADMM in some common cases (e.g., \textbf{Algorithm 2}) has been well studied in various practical cases \citep{zhou2016bilevel,xu2019nonlocal,yao2019nonconvex}, although a \emph{strictly} mathematical proof needs to be further improved and perfected. Moreover, we experimentally record the relative loss of the objective function in each iteration to draw the convergence curves of the proposed S2FL model on three datasets used in the experiment section, as shown in Fig. \ref{fig:Conv_curves}.

Furthermore, it is clear to observe that the overall computational cost of the optimization problem (\ref{eq1}) (i.e., our S2FL model) is mainly dominated by classic matrix algebra, such as matrix multiplication and matrix inversion. In detail, the update of the linear regression matrix $\mathbf{P}$ exhibits a total complexity with $\mathcal{O}(d_s^3+d_s^2KN+d_s^2C+d_sd_kK^2N+d_sCKN)$. For each iteration of \textbf{Algorithm 2} in solving the subproblem (\ref{eq6}), optimizing $\mathbf{\Theta_0}$ and $\mathbf{H}$ generally yields the costs of $\mathcal{O}(d_k^3K^3+d_k^2d_sK^2+d_k^2K^3N+d_kd_sK^2N+d_kK^3N^2)$ and $\mathcal{O}(d_s^3+d_s^2KN+d_s^2C+d_sd_kKN+d_sd_kCK+d_kK^2CN+d_sCKN)$, respectively, while updating the variable $\mathbf{G}$ requires computing a SVD with the order of cost as $\mathcal{O}(\min(d_k^2d_sK^2, d_s^2d_kK))$. The final update of the specific subspace projection matrices $\{\mathbf{\Theta}_{k}\}_{k=1}^{K}$ bears the same complexity as that of the \textbf{Algorithm 2}.

\begin{figure*}[!t]
	  \centering
			\includegraphics[width=1\textwidth]{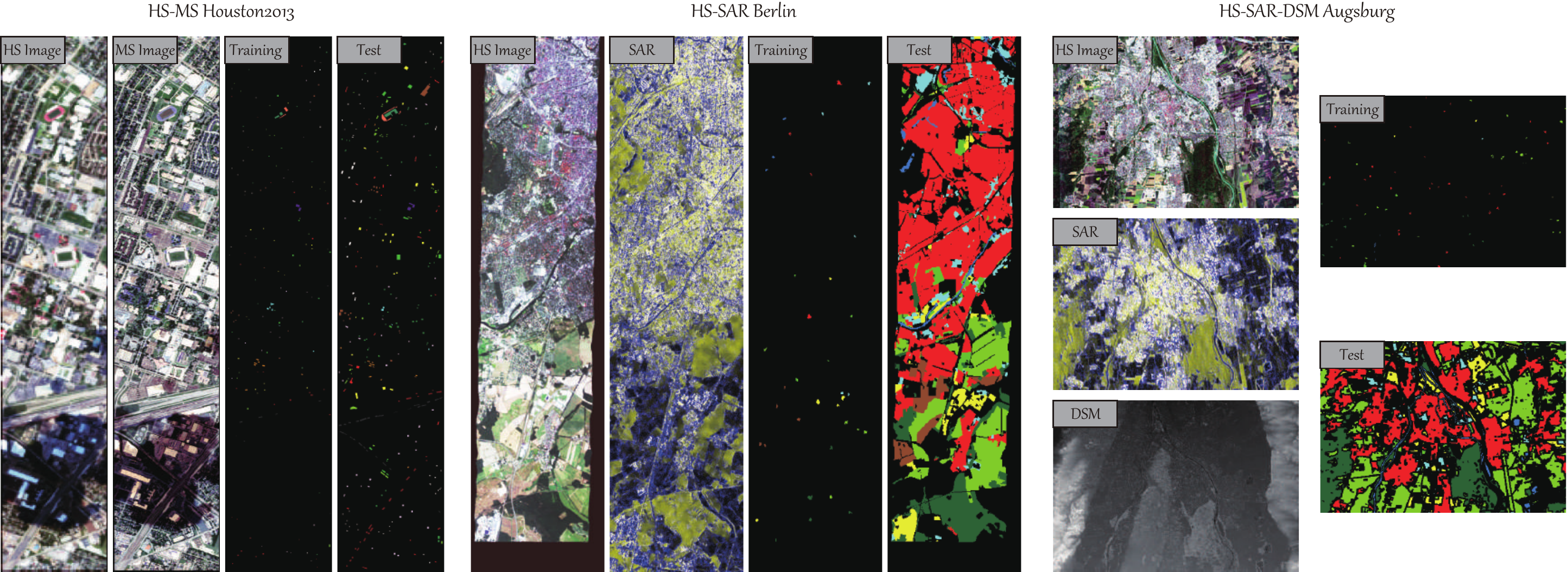}
        \caption{False-color visualization as well as the spatial distribution of training and test samples of three multimodal RS benchmark datasets: HS-MS \textit{Houston2013}, HS-SAR \textit{Berlin}, and HS-SAR-DSM \textit{Augsburg}.}
\label{fig:Datashow}
\end{figure*}

\section{Experiments}

\subsection{Multimodal Benchmark Datasets}

\subsubsection{HS-MS Houston2013 Data}
The scene consists of HS and MS data, which is a typical homogeneous dataset. The original HS image is available from IEEE GRSS data fusion contest 2013\footnote{\url{http://www.grss-ieee.org/community/technical-committees/data-fusion/2013-ieee-grss-data-fusion-contest/}} and has been widely concerned and applied for land cover classification. This image acquires the campus area of the University of Houston, Texas, USA, with $349\times1905$ pixels and $144$ channels covering the spectral range from $0.38 \mu m$ to $1.05 \mu m$. To make full use of high spectral information of the HS image and high spatial information of the MS image, we generate the HS-MS Houston2013 benchmark datasets by degrading the original HS image in spatial and spectral domains. More specifically,

\noindent \textbf{Spectral degradation:} The low spectral resolution MS image can be obtained by using the spectral response functions (SRFs) of the Sentinel-2 sensor. The resulting MS image is composed of the same size with the original HS image and $8$ spectral bands at a ground sampling distance (GSD) of $2.5m$.

\noindent \textbf{Spatial degradation:} The low spatial resolution HS image with a $10m$ GSD is generated by the means of the bilinear interpolation. To meet the pixel-to-pixel correspondences between HS and MS images, the degraded HS image is re-upsampled to the size of the MS image, i.e., $349\times1905$, by using the nearest neighbor interpolation.

Table \ref{Table:H2013} lists the types of ground objects and the number of training and test samples used for the classification task in the HS-MS Houston2013 datasets, and correspondingly Fig. \ref{fig:Datashow} visualizes the false-color images and the distribute of training and test sets.

\begin{table}[!t]
\centering
\caption{Description of the investigated HS-MS Houston2013 datasets, including the types of ground objects and the corresponding number of training and test samples.}
\vspace{2mm}
\resizebox{0.5\textwidth}{!}{ 
\begin{tabular}{c||ccc}
\toprule[1.5pt]
No. & Ground Object Name & Training Set & Test Set\\
\hline \hline 1 & Healthy Grass & 198 & 1053\\
 2 & Stressed Grass & 190 & 1064\\
 3 & Synthetic Grass & 192 & 505\\
 4 & Tree & 188 & 1056\\
 5 & Soil & 186 & 1056\\
 6 & Water & 182 & 143\\
 7 & Residential & 196 & 1072\\
 8 & Commercial & 191 & 1053\\
 9 & Road & 193 & 1059\\
 10 & Highway & 191 & 1036\\
 11 & Railway & 181 & 1054\\
 12 & Parking Lot1 & 192 & 1041\\
 13 & Parking Lot2 & 184 & 285\\
 14 & Tennis Court & 181 & 247\\
 15 & Running Track & 187 & 473\\
\hline \hline & Total & 2832 & 12197\\
\bottomrule[1.5pt]
\end{tabular}
}
\label{Table:H2013}
\end{table}

\begin{table}[!t]
\centering
\caption{Description of the investigated HS-SAR Berlin datasets, including the types of ground objects and the corresponding number of training and test samples.}
\vspace{2mm}
\resizebox{0.5\textwidth}{!}{ 
\begin{tabular}{c||ccc}
\toprule[1.5pt]
No. & Ground Object Name & Training Set & Test Set\\
\hline \hline 1 & Forest & 443 & 54511\\
 2 & Residential Area & 423 & 268219\\
 3 & Industrial Area & 499 & 19067\\
 4 & Low Plants & 376 & 58906\\
 5 & Soil & 331 & 17095\\
 6 & Allotment & 280 & 13025\\
 7 & Commercial Area & 298 & 24526\\
 8 & Water & 170 & 6502\\
\hline \hline & Total & 2820 & 461851\\
\bottomrule[1.5pt]
\end{tabular}
}
\label{Table:Berlin}
\end{table}

\subsubsection{HS-SAR Berlin Data}
The simulated EnMAP data synthesized based on the HyMap HS data graphically describes the Berlin urban and its rural neighboring area at $30m$ GSD, which can be freely downloaded from the website\footnote{{\url{http://doi.org/10.5880/enmap.2016.002}}}. In detail, there are $797\times220$ pixels in this scene, where 244 spectral bands are given in the wavelength range of $0.4 \mu m$ to $2.5 \mu m$. More details can be found in \citep{okujeni2016berlin}. To get a corresponding SAR data of the same region, we download a Sentinel-1 dual-Pol~(VV-VH) single look complex~(SLC) product from ESA. The product is collected under the Interferometric Wide (IW) swath mode. With the help of ESA toolbox SNAP\footnote{\url{https://step.esa.int/main/toolboxes/snap/}}, we build up a pre-processing workflow to prepare the SLC product as an analysis-ready SAR image. The workflow includes apply orbit profile, radiometric calibration, deburst, speckle reduction, terrain correction, and region-of-interest extraction. The analysis-ready SAR image used in this paper is geocoded in UTM/WGS84 coordinate system, and it is spatially averaged while applying speckle reduction and saved in $2\times 2$ PolSAR covariance matrix. Since the azimuth resolution of the Sentinel-1 data is approximately $13m$, the processed SAR image has a $13.89m$ GSD and consists of $1723\times476$ pixels. Similarly to the first datasets, the nearest neighbor interpolation is performed on the HS image, enabling the same image size with the SAR data.

The ground reference data for land cover classification is generated by using the OpenStreetMap data~\citep{haklay2008openstreetmap}, where the information regarding the training and test sets is shown in Fig. \ref{fig:Datashow} and Table \ref{Table:Berlin}. 

\subsubsection{HS-SAR-DSM Augsburg Data}
This dataset consists of three different data sources, including a spaceborne HS image, a dual-Pol PolSAR image, and a DSM image, where HS and DSM data are acquired by DAS-EOC, DLR, and the PolSAR data are collected from the Sentinel-1 platform, over the city of Augsburg, Germany. They are collected by the HySpex sensor \citep{baumgartner2012characterisation}, the Sentinel-1 sensor, and the DLR-3K system \citep{kurz2011real}, respectively. To evaluate the performance of multimodal fusion classification effectively, we downsample the spatial resolution of all images to a unified $30m$ GSD. The scene comprises of $332\times 485$ pixels and $180$ spectral bands ranging from $0.4 \mu m$ to $2.5 \mu m$ for the HS image, $1$ band for the DSM image, and $4$ features from the dual-Pol (VV-VH) SAR image. Note that the SAR data is preprocessed in the same way as the SAR component in HS-SAR Berlin Data. The $4$ features are VV intensity, VH intensity, the real part and the imaginary part of the off-diagonal element of the PolSAR covariance matrix. The ground reference data is generated from the OpenStreetMap data. Detailed information regarding the training and test sets is shown in Fig. \ref{fig:Datashow} and Table \ref{Table:augsburg}. 

\begin{table}[!t]
\centering
\caption{Description of the investigated HS-SAR Augsburg datasets, including the types of ground objects and the corresponding number of training and test samples.}
\vspace{2mm}
\resizebox{0.5\textwidth}{!}{ 
\begin{tabular}{c||ccc}
\toprule[1.5pt]
No. & Ground Object Name & Training Set & Test Set\\
\hline \hline 1 & Forest & 146 & 13361\\
 2 & Residential Area & 264 & 30065\\
 3 & Industrial Area & 21 & 3830\\
 4 & Low Plants & 248 & 26609\\
 5 & Allotment & 52 & 523\\
 6 & Commercial Area & 7 & 1638\\
 7 & Water & 23 & 1507\\
\hline \hline & Total & 761 & 77533\\
\bottomrule[1.5pt]
\end{tabular}
}
\label{Table:augsburg}
\end{table}

\subsection{Experimental Setup and Preparation}

\subsubsection{Evaluation Metrics}
In the experiments, land cover classification is regarded as a potential application to evaluate the quality of the multimodal feature representations learned by the proposed S2FL model. More specifically, we compute three widely-used criteria, i.e., \textit{Overall Accuracy (OA)}, \textit{Average Accuracy (AA)}, and \textit{Kappa Coefficient} ($\kappa$) to make a quantitative performance comparison using a nearest neighbor (NN) classifier. More specifically, we first apply the proposed S2FL model to extract or learn the multimodal feature representations and then feed them into a classifier (NN in our case). The main reason to select the NN classifier can be clarified by the fact that we expect to see the performance gain owing to the learned features obtained by our proposed method rather than those advanced classifiers, e.g., support vector machine (SVM), random forest (RF), deep learning-based classifiers, etc. 

\subsubsection{Comparison with State-of-the-art MFL Models}
Several state-of-the-art MFL algorithms related to land cover classification of multimodal RS images are selected as competitors, compared to the proposed S2FL model. They are joint dimensionality reduction based on principal components analysis (JDR-PCA)\citep{martinez2001pca}, supervised manifold alignment (SMA) \citep{wang2011heterogeneous}, unsupervised manifold alignment (USMA) \citep{liao2016manifold}, common subspace learning ($\ell_{2}$-CoSpace) \citep{hong2019cospace}, $\ell_{1}$-CoSpace \citep{hong2020learning}, as well as single modalities, e.g., HS, MS, SAR, DSM, and their simple concatenation, e.g., HS+MS, HS+SAR, HS+SAR+DSM.

\subsubsection{Implementation Details}
The algorithm performance, to a great extent, depends on the parameter tuning, e.g., regularization parameters in CoSpace and our S2FL. As a result, selecting a proper range for these parameters is of great significance in practical applications. For this reason, a 10-fold cross validation is conducted on the training set to determine the parameter combination of different methods. These parameters are tuned in a given range to maximize the classification accuracy. For example, the number of nearest neighbors ($q$) and the parameter $\sigma$ for computing the adjacency matrix $\mathbf{W}$ in USMA and S2FL can be selected from $\{5, 10,...,50\}$ and $\{10^{-2}, 10^{-1}, 10^{0}, 10^{1}, 10^{2}\}$, respectively. The regularization parameters, e.g., $\alpha$ and $\beta$ in CoSpace and S2FL, can be determined in the range of $\{10^{-3},10^{-2}, 10^{-1}, 10^{0}, 10^{1}, 10^{2}\}$. Moreover, the feature dimension is also a key parameter, which has great effects on the quality of final learned feature representations. For that, the optimal feature dimension ($d$) can be found out from $5$ to $50$ at a $5$ interval, according to the best classification performance on the training set.

It should be noted, however, that the DSM image only holds one feature band in the \textit{Augsburg} datasets, hence attribute profiles are first extracted to fully utilize the spatial information of the DSM image, extending the feature bands to $3$, before performing feature learning and classification. 

\begin{table*}[!t]
\centering
\caption{Quantitative results of different compared approaches in terms of OA, AA, and $\kappa$ as well as the accuracy for each class on HS-MS \textit{Houston2013} datasets using NN classifier, where the parameters are determined by cross-validation on the training sets. The best one is shown in bold.}
\vspace{2mm}
\resizebox{0.92\textwidth}{!}{ 
\begin{tabular}{c||ccccccccc}
\toprule[1.5pt] Method & MS & HS & HS+MS & JDR-PCA & USMA & SMA & $\ell_{2}$-CoSpace & $\ell_{1}$-CoSpace & S2FL\\
\hline \hline
\multirow{2}{*}{Parameters} & \multirow{2}{*}{--} & \multirow{2}{*}{--} & \multirow{2}{*}{--} & $d$ & $q,\sigma, d$ & $d$ & $\alpha, \beta, d$ & $\alpha, \beta, d$ & $\alpha, \beta, d$\\
& & & & $20$ & $10, 1, 30$ & $30$ & $1, 0.01, 30$ & $0.1, 0.1, 30$ & $0.01, 0.1, 30$\\
\hline \hline
Healthy Grass & \bf 82.53 & 76.64 & 79.39 & 79.39 & 82.05 & 81.01 & 81.10 & 80.15 & 80.06\\
Stressed Grass & 81.95 & 74.25 & 76.60 & 76.60 & 80.83 & \bf 84.68 & 83.46 & 83.36 & 84.49\\
Synthetic Grass & 98.81 & 91.09 & 93.47 & 93.07 & 99.01 & 97.82 & 97.03 & \bf 99.41 & 98.02 \\
Trees & \bf 91.29 & 70.17 & 81.72 & 81.72 & 90.81 & 88.45 & 88.54 & 89.77 & 87.31\\
Soil & 96.31 & 98.48 & 98.48 & 98.48 & 94.89 & 97.25 & 98.39 & 99.72 & \bf 100.00\\
Water & \bf 98.60 & 77.62 & 81.12 & 81.12 & 95.10 & 83.22 & 95.10 & 81.82 & 83.22\\
Residential & 72.01 & 67.07 & 73.51 & 73.41 & 70.34 & 74.72 & \bf 77.99 & 71.08 & 73.32 \\
Commercial & 34.00 & 41.22 & 41.31 & 41.31 & 34.28 & 43.11 & 44.35 & 64.10 & \bf 74.84\\
Road & 63.74 & 60.15 & 63.36 & 63.36 & 67.23 & 62.13 & 66.19 & 65.91 & \bf 78.38\\
Highway & 43.44 & 44.79 & 46.14 & 46.14 & 41.80 & 40.64 & 58.01 & 60.81 & \bf 83.30\\
Railway & 64.52 & 57.12 & 57.50 & 57.50 & 57.21 & 51.71 & 70.59 & 79.70 & \bf 81.69\\
Parking Lot 1 & 46.88 & 79.54 & 75.50 & 75.41 & 45.24 & 62.73 & 87.80 & 83.57 & \bf 95.10 \\
Parking Lot 2 & 52.28 & 70.53 & \bf 73.33 & \bf 73.33 & 38.25 & 60.00 & 67.02 & 70.18 & 72.63 \\
Tennis Court & 96.36 & 88.26 & 97.57 & 97.57 & 95.55 & 98.79 & 98.79 & \bf 100.00 & \bf 100.00 \\
Running Track & 98.52 & 82.24 & 90.49 & 90.49 & 97.89 & 98.31 & \bf 99.37 & 98.94 & \bf 99.37\\
\hline \hline
OA (\%) & 70.82 & 69.21 & 72.02 & 71.98 & 69.39 & 71.66 & 77.97 & 79.86 & \bf 85.07\\
AA (\%) & 74.75	& 71.94 & 75.30 & 75.26 & 72.70 & 74.97 & 80.91 & 81.90 & \bf 86.11\\
$\kappa$ & 0.6866 & 0.6684 & 0.6984 & 0.6980 & 0.6705 & 0.6929 & 0.7629 & 0.7814 & \bf 0.8378\\
\bottomrule[1.5pt]
\end{tabular}}
\label{tab:H2013}
\end{table*}

\begin{figure*}[!t]
	  \centering
			\includegraphics[width=0.9\textwidth]{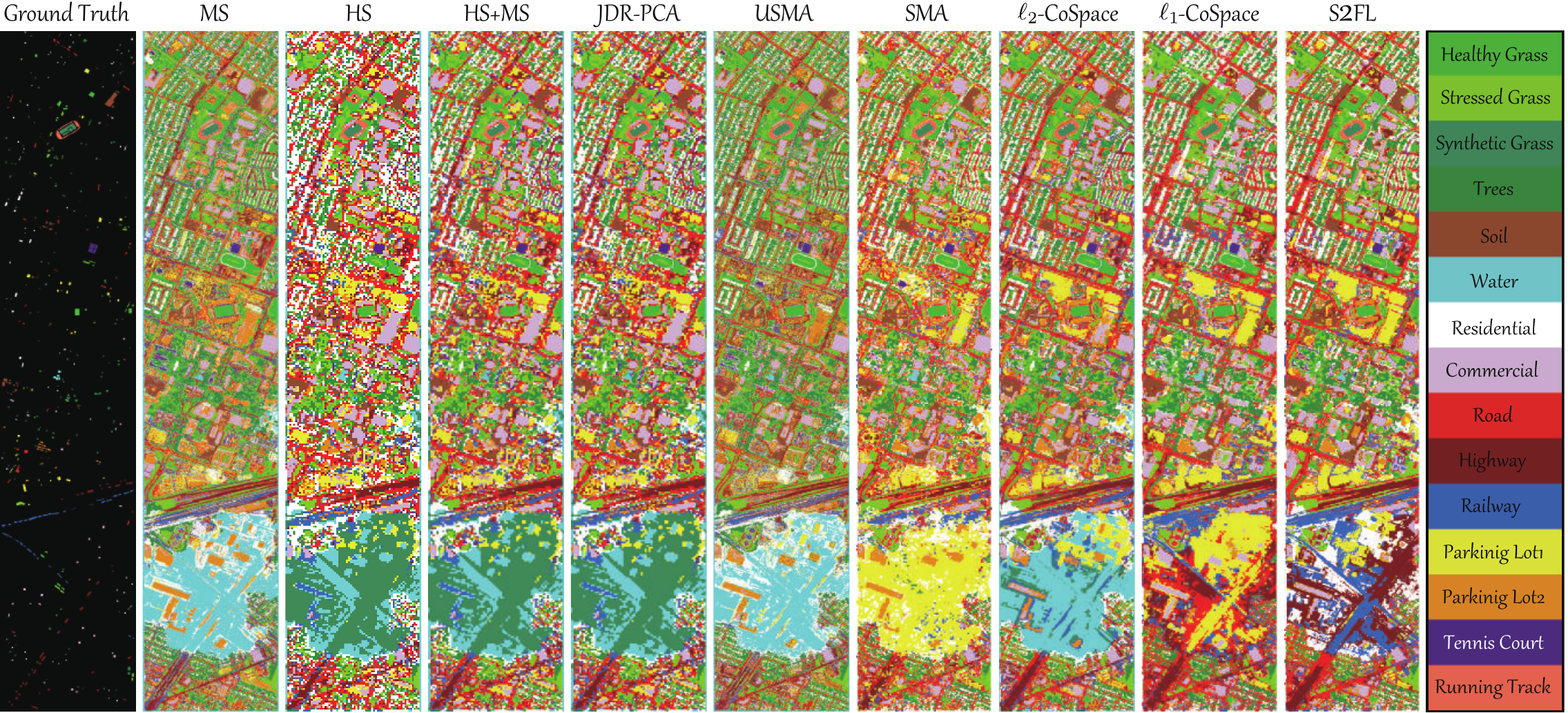}
        \caption{Classification maps obtained by different MFL algorithms on the \textit{Houston2013} datasets.}
\label{fig:CM_HH_IF}
\end{figure*}

\subsection{Results and Analysis on Houston2013 Datasets}

Table \ref{tab:H2013} lists the quantitative classification results of compared feature learning methods on HS-MS \textit{Houston2013} datasets, including \textit{OA}, \textit{AA}, $\kappa$, and the accuracy for each class. Overall, the joint use of multiple modalities (e.g., HS+MS) obviously performs better than single modalities in three main indices, i.e., \textit{OA}, \textit{AA}, and $\kappa$. The classification accuracies obtained by JDR-PCA are basically same to those jointly using HS and MS data. For MA-based approaches (e.g., USMA, SMA), they fail to classify the materials well, due to their sensitivity to various complex noises. This indirectly leads to relatively poor performance, compared to PCA and HS+MS. Owing to the use of label (supervised) information, the classification accuracy of SMA is higher than that of USMA, bringing an increase of about 2\% points OA. Moreover, the joint learning-based group, e.g., $\ell_{2}$-CoSpace, $\ell_{1}$-CoSpace, and S2FL, dramatically outperforms other competitors, either in main indices (\textit{OA}, \textit{AA}, and $\kappa$) or in the accuracy for the majority of categories. The performance of $\ell_{1}$-CoSpace is superior to that of $\ell_{2}$-CoSpace (approximately 2\% points improvement), since the feature selection strategy is performed in $\ell_{1}$-CoSpace by the means of sparsity-promoting $\ell_{1}$-norm term. More remarkably, our proposed S2FL model can obtain a higher classification result with around 5\% points gain in \textit{OA} on the basis of $\ell_{1}$-CoSpace that ranks the second place. In addition, S2FL also plays a dominated role in the classification accuracy of each class. That is, S2FL is capable of achieving the best classification performance in many categories (e.g., \textit{Soil}, \textit{Commercial}, \textit{Road}, \textit{Highway}, \textit{Railway}, \textit{Tennis Court}, and \textit{Running Track}), which demonstrates its effectiveness and superiority in the land cover classification task.

Visually, Fig. \ref{fig:CM_HH_IF} shows the integral classification maps of considered compared algorithms in the given scene. There is a basically identical trend in both quantitative and qualitative comparisons (between Fig. \ref{fig:CM_HH_IF} and Table \ref{tab:H2013}). It is worth noting that the classification map obtained by the proposed S2FL model has not only more structurally geometric information for those man-made materials, e.g., \textit{Residential}, \textit{Commercial}, \textit{Parking Lot}, etc., but also more detailed textural information, e.g., for the ground objects of \textit{Grass} and \textit{Tree}.

\subsubsection{Parameter Analysis of S2FL Model}
\begin{figure*}[!t]
	  \centering
		\subfigure{
			\includegraphics[width=0.305\textwidth]{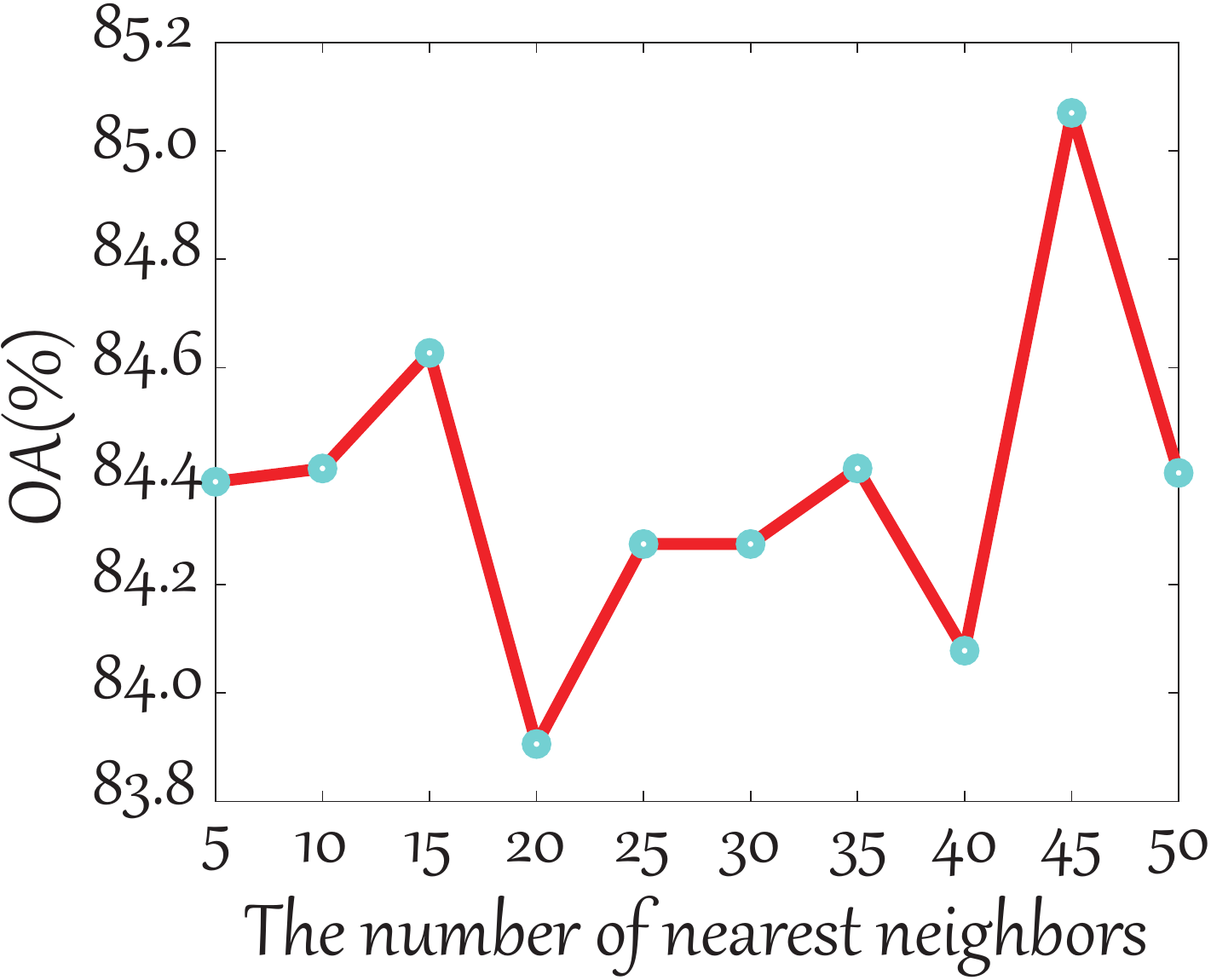}
		}
		\subfigure{
			\includegraphics[width=0.305\textwidth]{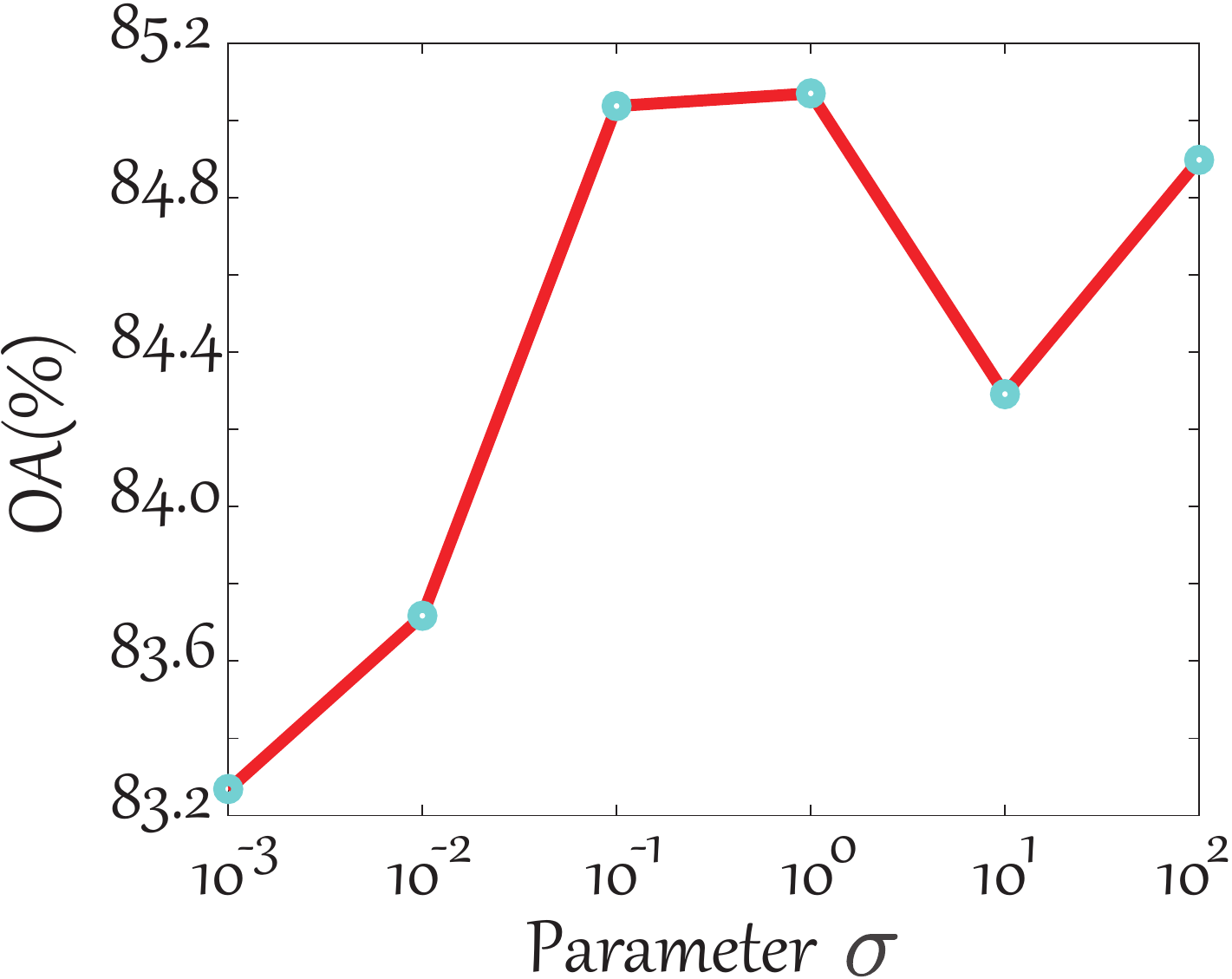}
		}
		\subfigure{
			\includegraphics[width=0.305\textwidth]{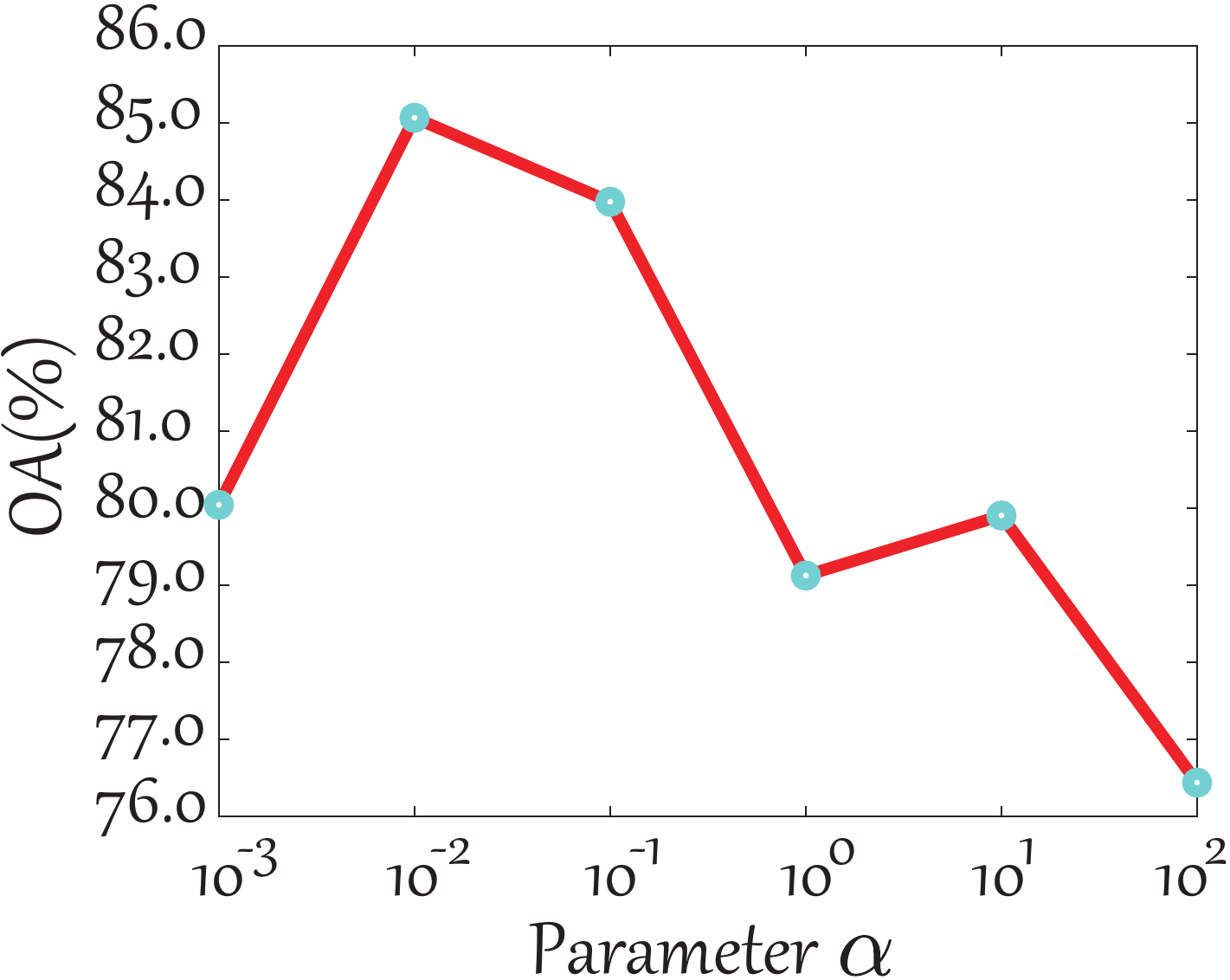}
		}
		\subfigure{
			\includegraphics[width=0.305\textwidth]{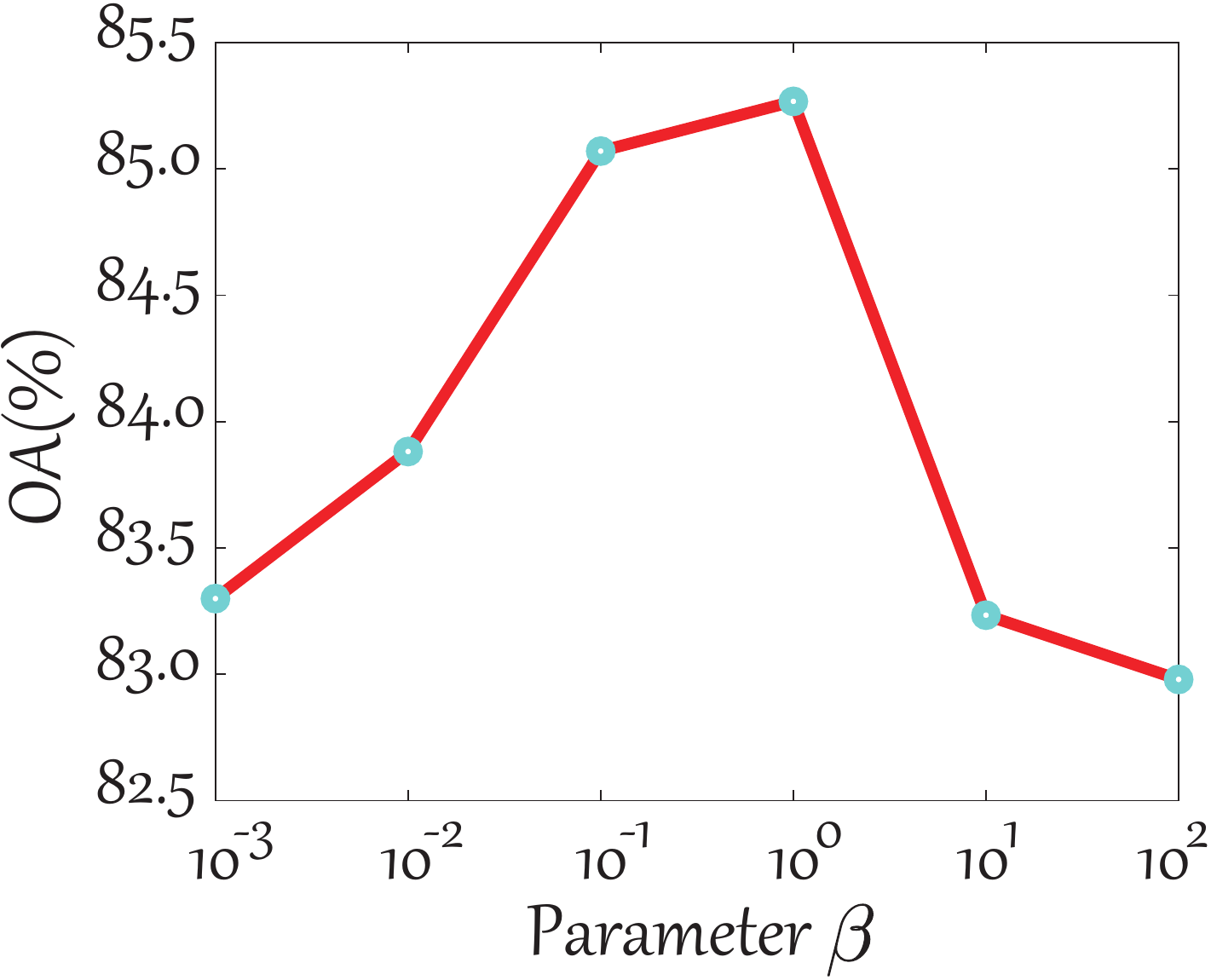}
		}
		\subfigure{
			\includegraphics[width=0.305\textwidth]{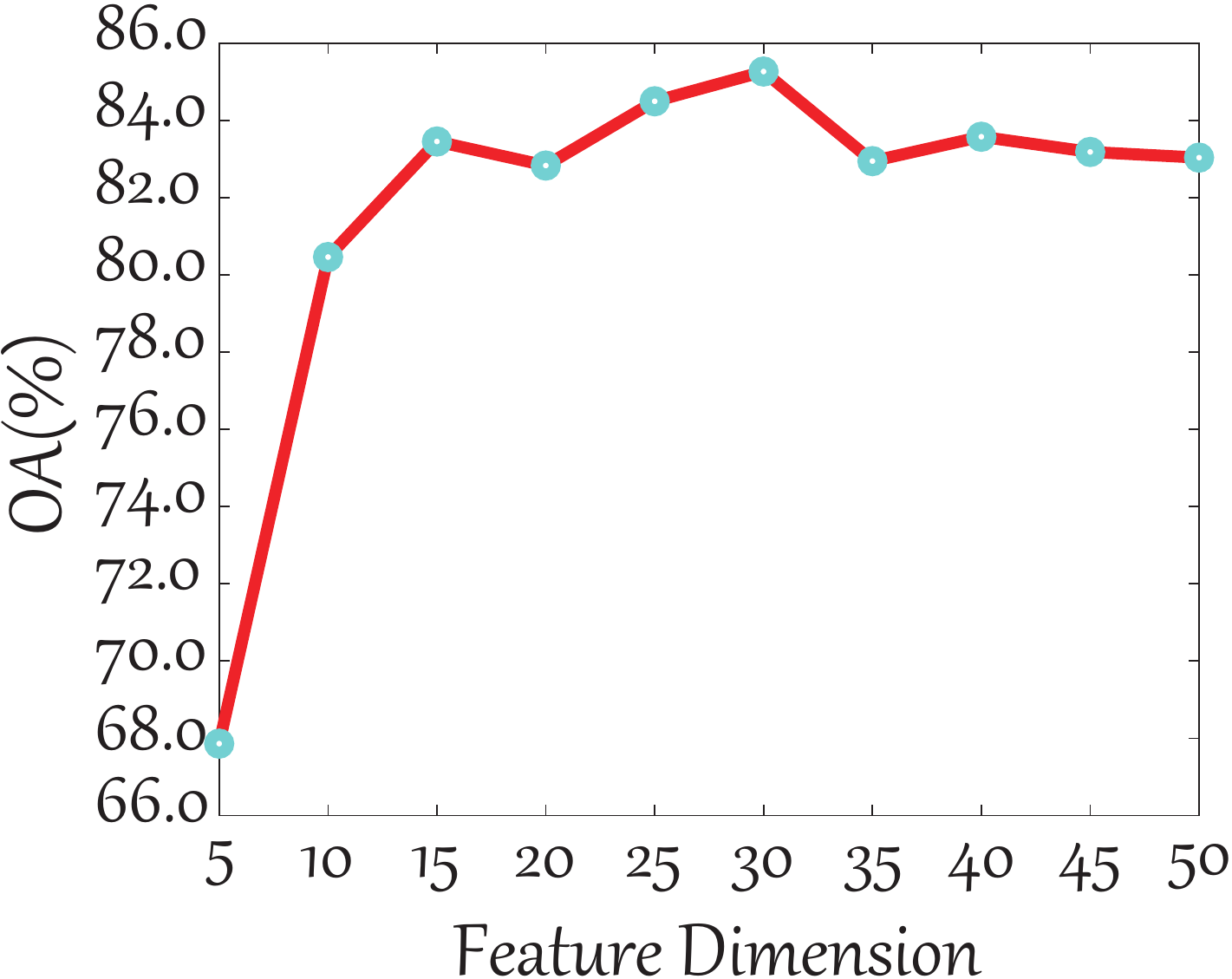}
		}
\caption{Parameter sensitivity analysis of S2FL model conducted on the HS-MS \textit{Houston2013} datasets, e.g., the number of nearest neighbors and $\sigma$ in Eq. (\ref{eq2}), the regularization parameters $\alpha$ and $\beta$ in Eq. (\ref{eq1}), and the feature dimension.}
\label{fig:parameters}
\end{figure*}

Parameter (or model) selection is the key factor to wield significant influence over the performance gain of the feature learning method. It is necessary, therefore, to discuss and analyze the parameter sensitivity. There are five main parameters, i.e., the number of nearest neighbors ($q$) and the parameter $\sigma$ in Eq. (\ref{eq2}), the regularization parameters ($\alpha$ and $\beta$) in Eq. (\ref{eq1}), and the feature dimension ($d$), in the S2FL model. As shown in Fig. \ref{fig:parameters}, the regularization parameter $\alpha$ and the feature dimension $d$ yield a higher change in the term of \textit{OA} with different input values, compared to other three parameters. Moreover, varying the number of nearest neighbors $q$ is relatively insensitive to the classification accuracy, while the parameters $\sigma$ and $\beta$ can moderately control the change of \textit{OA}. From the overall perspectives, the \textit{OA} value remains stable as long as these parameters are selected in a proper range. The optimal parameter combination is $(q, \sigma, \alpha, \beta, d)=(45, 10^{0}, 10^{-2}, 10^{0}, 30)$, which is basically identical to that obtained by cross-validation on the training set. This, to some extent, demonstrates the acceptable practicability of the proposed S2FL model in the real application.

\begin{table*}[!t]
\centering
\caption{Quantitative results of different compared approaches in terms of OA, AA, and $\kappa$ as well as the accuracy for each class on HS-SAR \textit{Berlin} datasets using NN classifier, where the parameters are determined by cross-validation on the training sets. The best one is shown in bold.}
\vspace{2mm}
\resizebox{0.9\textwidth}{!}{ 
\begin{tabular}{c||ccccccccc}
\toprule[1.5pt] Method & SAR & HS & HS+SAR & JDR-PCA & USMA & SMA & $\ell_{2}$-CoSpace & $\ell_{1}$-CoSpace & S2FL\\
\hline \hline
\multirow{2}{*}{Parameters} & \multirow{2}{*}{--} & \multirow{2}{*}{--} & \multirow{2}{*}{--} & $d$ & $q,\sigma, d$ & $d$ & $\alpha, \beta, d$ & $\alpha, \beta, d$ & $\alpha, \beta, d$\\
& & & & $10$ & $10, 1, 10$ & $10$ & $1, 0.01, 10$ & $0.1, 1, 10$ & $0.01, 1, 10$\\
\hline \hline
Forest & 24.47 & 71.45 & 73.57 & 73.56 & 62.19 & 57.34 & 79.70 & 82.94 & \bf 83.30\\
Residential Area & 19.86 & 41.19 & 42.93 & 42.81 & 43.29 & 50.53 & 50.51 & 52.75 & \bf 57.39\\
Industrial Area & 25.53 & 40.37 & 38.72 & 38.66 & 27.97 & 43.46 & \bf 49.63 & 44.71 & 48.53\\
Low Plants & 31.79 & 68.07 & 69.23 & 68.03 & 63.46 & 50.03 & 69.27 & 76.54 & \bf 77.16\\
Soil & 31.61 & 81.05 & 81.01 & 81.01 & 79.02 & 73.60 & 70.30 & 81.92 & \bf 83.84\\
Allotment & 13.32 & 63.89 & 62.89 & 61.90 & 48.33 & 52.69 & \bf 66.23 & 55.36 & 57.05\\
Commercial Area & 14.96 & 35.31 & 36.83 & \bf 36.94 & 23.51 & 26.59 & 32.02 & 29.64 & 31.02\\
Water & 7.91 & 65.76 & 67.13 & \bf 67.21 & 47.34 & 65.20 & 61.80 & 45.36 & 61.57\\
\hline \hline
OA (\%) & 21.98 & 50.30 & 51.71 & 51.47 & 47.93 & 50.83 & 56.67 & 58.84 & \bf 62.23\\
AA (\%) & 21.18 & 58.39 & 59.04 & 58.77 & 49.39 & 52.43 & 59.93 & 58.65 & \bf 62.48\\
$\kappa$ & 0.0759 & 0.3709 & 0.3838 & 0.3809 & 0.3243 & 0.3528 & 0.4306 & 0.4571 & \bf 0.4877\\
\bottomrule[1.5pt]
\end{tabular}
}
\label{tab:Be}
\end{table*}

\begin{table*}[!t]
\centering
\caption{Ablation analysis of the proposed S2FL model on the Houston2013 datasets.}
\vspace{2mm}
\resizebox{0.8\textwidth}{!}{ 
\begin{tabular}{c||ccccc|ccc}
\toprule[1.5pt] Model & HS & MS & Orthogonality & Shared & Specific & OA (\%) & AA (\%) & $\kappa$\\
\hline \hline
-- & \cmark & \xmark & \xmark & \xmark & \xmark & 69.21 & 71.94 & 0.6684\\
-- & \xmark & \cmark & \xmark & \xmark & \xmark & 70.82 & 74.75 & 0.6866\\
S2FL & \cmark & \xmark & \cmark & \xmark & \cmark & 76.50 & 78.80 & 0.7456\\
S2FL & \xmark & \cmark & \cmark & \xmark & \cmark & 72.74 & 76.38 & 0.7070\\
S2FL & \cmark & \cmark & \cmark & \cmark & \xmark & 78.77 & 81.59 & 0.7697\\
S2FL & \cmark & \cmark & \cmark & \xmark & \cmark & 83.11 & 84.52 & 0.8166\\
S2FL & \cmark & \cmark & \xmark & \cmark & \cmark & 66.11 & 69.82 & 0.6321\\
S2FL & \cmark & \cmark & \cmark & \cmark & \cmark & \bf 85.07 & \bf 86.01 & \bf 0.8378\\
\bottomrule[1.5pt]
\end{tabular}
}
\label{tab:AB}
\end{table*}

\subsection{Ablation Analysis of the Proposed S2FL Model}
To verify the effectiveness of the proposed idea (i.e., shared and specific feature disentangling) in the S2FL model, we investigate the performance gain with the use of different components, i.e., only using modality-shared features (obtained via $\mathbf{\Theta}_{0}$), only using modality-specific features (obtained via $\mathbf{\Theta}_{k}$). The quantitative results in terms of \textit{OA}, \textit{AA}, and $\kappa$ indices on the \textit{Houston2013} datasets are reported in Table \ref{tab:AB}. More specifically, the S2FL's results only using one modality (e.g., HS, MS)\footnote{In this case, our S2FL is reduced to the single modality feature learning model, i.e., only one $\mathbf{\Theta}_{k}$ for either HS or MS.} are reported, which is better than those without feature learning (the first two rows in Table \ref{tab:AB}). We also consider the case of only using shared features or only using specific features in our S2FL model for land cover classification. As can be seen from Table \ref{tab:AB}, the performance of S2FL by considering multimodal input is superior to that of only using single modalities, while the modality-specific information is more important than the modality-shared information (\textit{OA}: 83.11 vs 78.77) in the classification task. Moreover, we found that the performance happens a dramatic degradation without the orthogonal constraint in the S2FL model. Not unexpectedly, our S2FL with a joint combination of shared and specific features achieves the best results.

\subsection{Cross-modality Experiments: A Special Case of Multi-modality}
Take the bi-modality as an example, cross-modality learning (CML) for simplicity refers to that training a model using two modalities and one modality is absent in the testing phase, or \emph{vice versa} (only one modality is available for training and bi-modality for testing) \citep{ngiam2011multimodal}. Such a CML problem that exists widely in various RS tasks is more applicable to real-world cases. n recent years, there have been some works proposed to investigate the CML issue and applied for land cover classification. More details regarding the CML's setting can refer to \citep{hong2019cospace,hong2019learnable}. Table \ref{tab:CML} lists the quantitative comparison between the single modalities and the CML's cases using our S2FL. There is a basically consistent trend in both HS and MS data. That is, the classification accuracies of directly using the original spectral features (either HS or MS) are lower than those using learned features (e.g., via S2FL). In addition, the features learned by the S2FL-CML setting can better classify the land cover materials compared to those directly learning features from the single modalities (e.g., S2FL-HS or S2FL-MS), showing the effectiveness of the proposed S2FL model for both cases of multi-modality learning and the CML.

\begin{table*}[!t]
\centering
\caption{Quantitative comparison using the S2FL model under the cross-modality setting. ``Only HS (MS)'' means directly using original HS (MS) data as the features, ``S2FL-HS (MS)'' means the single modality feature learning (e.g., HS or MS), and ``S2FL-CML-HS (MS)'' means the CML setting, i.e., training on HS-MS data and only HS (MS) data are available in the testing phase.}
\vspace{2mm}
\resizebox{0.85\textwidth}{!}{ 
\begin{tabular}{c||ccc|ccc}
\toprule[1.5pt] Method & Only HS & S2FL-HS & S2FL-CML-HS & Only MS & S2FL-MS & S2FL-CML-MS\\
\hline \hline
OA (\%) & 69.21 & 76.50 & 80.86 & 70.01 & 72.74 & 77.01\\
AA (\%) & 71.94 & 78.80 & 82.21 & 80.46 & 76.38 & 80.46\\
$\kappa$ & 0.6684 & 0.7456 & 0.7922 & 75.09 & 0.7070 & 0.7509\\
\bottomrule[1.5pt]
\end{tabular}
}
\label{tab:CML}
\end{table*}

\begin{figure*}[!t]
	  \centering
			\includegraphics[width=1\textwidth]{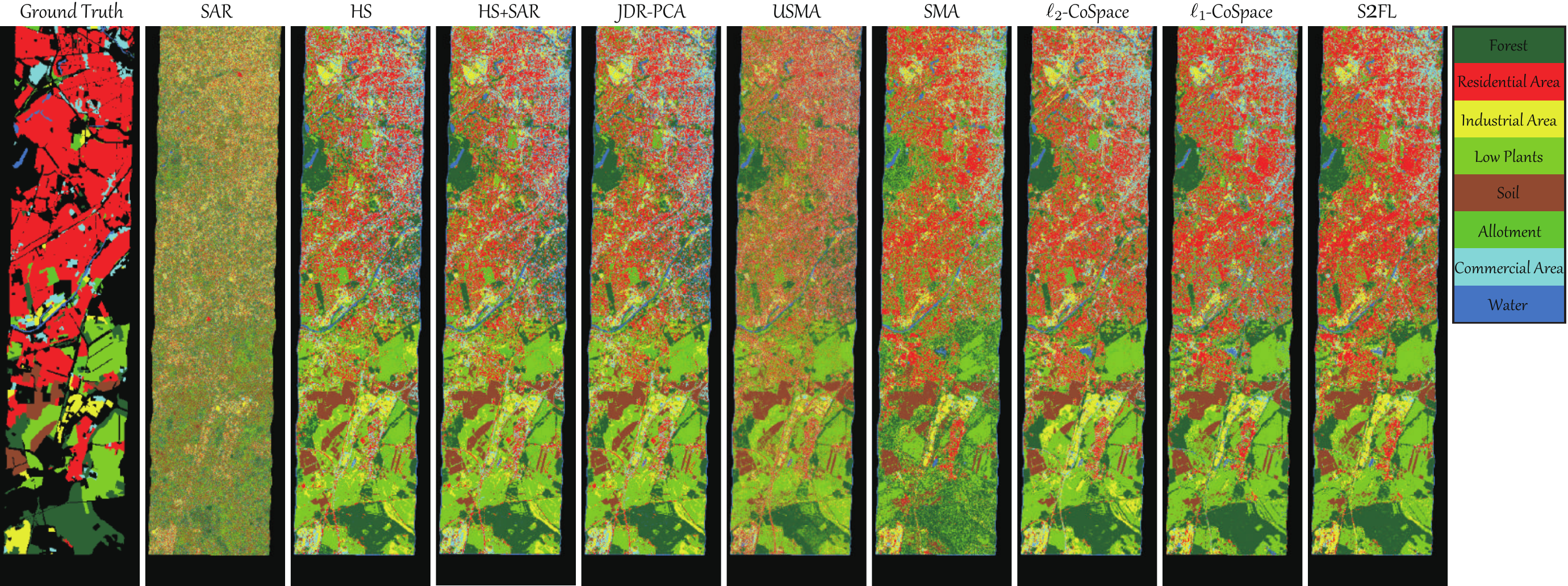}
        \caption{Classification maps obtained by different MFL algorithms on the \textit{Berlin} datasets.}
\label{fig:CM_Be_IF}
\end{figure*}

\subsection{Results and Analysis on Berlin Datasets}
Unlike the homogeneous HS-MS data, the heterogeneity between HS and SAR data remains challenging in the feature learning and fusion. Therefore, the quantitative comparison conducted on the \textit{Berlin} datasets (see Table \ref{tab:Be}) becomes significantly meaningful for the performance assessment of MFL models. As can be seen in Table \ref{tab:Be}, the heterogeneous datasets are challenging and difficult to perform the land cover classification, yielding a sharp decrease in classification performance compared to results on the \textit{Houston2013} datasets. Nevertheless, the whole trend between different algorithms is similar. The classification accuracy using the concatenation of HS and SAR data is still higher than that only using single modalities. PCA-based joint feature learning obtains a basically same result with HS+SAR's. Similarly, USMA and SMA fail to align the heterogeneous data well. The reasons are two-fold: the sensitivity to the noises and only considering the shared component representations across modalities. The CoSpace family, i.e., $\ell_{2}$-CoSpace and $\ell_{1}$-CoSpace, is robust to the noise effects by learning the latent subspace to bride the multimodal data and label information, bringing increments of 5.84\% and 8.01\% points \textit{OA}, respectively, on the basis of SMA. Not unexpectedly, our S2FL method achieves the best performance with further improvement of 3.39\%, 3.83\%, and 3.06\% points with respect to \textit{OA}, \textit{AA}, and $\kappa$ over $\ell_{1}$-CoSpace. Additionally, the S2FL model can also obtain desirable classification accuracy for each class in comparison with other approaches.

Fig. \ref{fig:CM_Be_IF} shares a similar visual comparison with quantitative results. Only using SAR data yields a poor classification map with extensive noisy points. Notably, our method is capable of fully blending the HS and SAR information by the means of the interpretable shared and specific feature learning mechanism, thereby reducing the noisy pixels and generating more smooth classification maps. Particularly for those ground objects that hold rich texture information, e.g., \textit{Forest}, \textit{Low Plants}, the S2FL model tends to capture their subtle differences against noises by the means of specific information of each modality. Furthermore, the learned common components can depict the structural information, thereby further identifying the materials of \textit{Residential} and \textit{Commercial} effectively.

\begin{figure*}[!t]
	  \centering
			\includegraphics[width=1\textwidth]{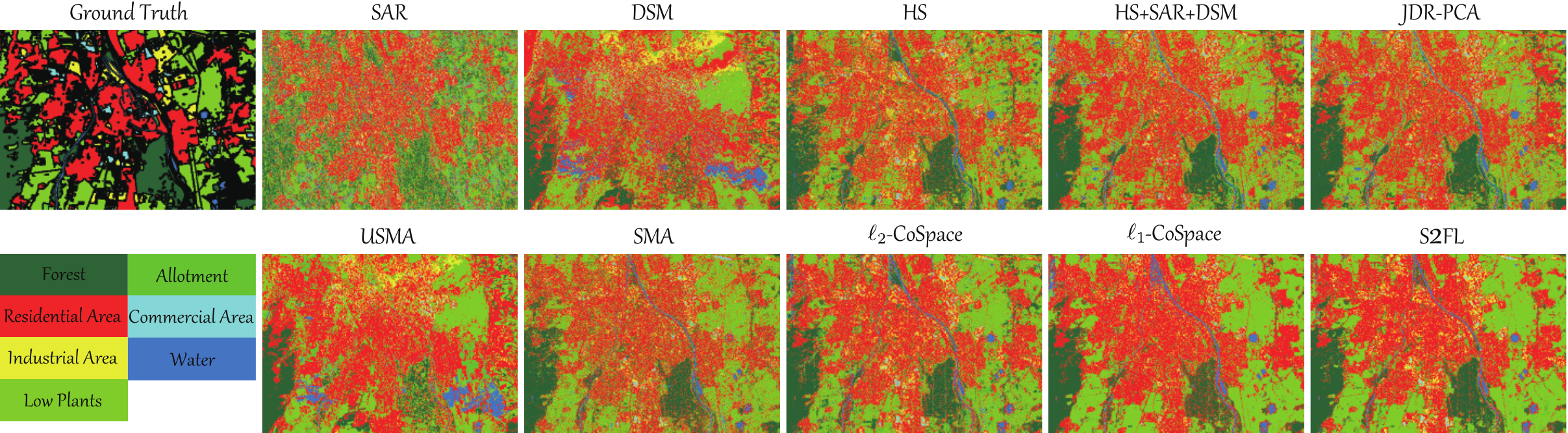}
        \caption{Classification maps obtained by different MFL algorithms on the \textit{Augsburg} datasets.}
\label{fig:CM_Au_IF}
\end{figure*}

\begin{table*}[!t]
\centering
\caption{Quantitative results of different compared approaches in terms of OA, AA, and $\kappa$ as well as the accuracy for each class on HS-SAR-DSM \textit{Augsburg} datasets using NN classifier, where the parameters are determined by cross-validation on the training sets. The best one is shown in bold.}
\vspace{2mm}
\resizebox{1\textwidth}{!}{ 
\begin{tabular}{c||cccccccccc}
\toprule[1.5pt] Method & SAR & DSM & HS & HS+SAR+DSM & JDR-PCA & USMA & SMA & $\ell_{2}$-CoSpace & $\ell_{1}$-CoSpace & S2FL\\
\hline \hline
\multirow{2}{*}{Parameters} & \multirow{2}{*}{--} & \multirow{2}{*}{--} & \multirow{2}{*}{--} & \multirow{2}{*}{--} & $d$ & $q,\sigma, d$ & $d$ & $\alpha, \beta, d$ & $\alpha, \beta, d$ & $\alpha, \beta, d$\\
& & & & & $20$ & $10, 1, 20$ & $20$ & $1, 0.1, 20$ & $0.1, 0.1, 20$ & $0.1, 0.01, 20$\\
\hline \hline
Forest & 44.40 & 53.18 & 81.75 & 88.39 & 88.32 & 65.53 & 81.94 & 82.04 & 87.79 & \bf 88.80\\
Residential Area & 70.10 & 66.70 & 77.01 & 82.91 & 82.93 & 79.14 & 73.43 & 83.52 & \bf 88.32 & 86.36\\
Industrial Area & 13.79 & 11.83 & 26.84 & 22.38 & 22.27 & 21.85 & 24.18 & 34.99 & 32.64 & \bf 38.90\\
Low Plants & 61.55 & 55.06 & 67.67 & 69.24 & 69.06 & 56.36 & 81.12 & 76.67 & 88.44 & \bf 90.53\\
Allotment & 9.18 & 25.24 & 46.65 & 55.45 & 55.07 & 44.17 & 22.75 & 52.77 & 45.89 & \bf 68.64\\
Commercial Area & 8.30 & 3.48 & 12.21 & 11.29 & 11.29 & 5.62 & 8.79 & \bf 14.22 & 12.09 & 8.97\\
Water & 6.97 & 5.84 & 43.20 & 47.38 & 47.38 & 17.98 & 32.85 & \bf 48.71 & 17.65 & 47.45\\
\hline \hline
OA (\%) & 57.01 & 54.87 & 69.91 & 73.78 & 73.71 & 63.17 & 72.61 & 76.17 & 81.49 & \bf 83.36\\
AA (\%) & 30.61 & 31.62 & 50.76 & 53.86 & 53.76 & 41.52 & 46.44 & 56.13 & 53.56 & \bf 61.38\\
$\kappa$ & 0.3974 & 0.3571 & 0.5742 & 0.6252 & 0.6241 & 0.4776 & 0.6046 & 0.6643 & 0.7327 & \bf 0.7626\\
\bottomrule[1.5pt]
\end{tabular}
}
\label{tab:Au}
\end{table*}

\subsection{Results and Analysis on Augsburg Datasets}
We further investigate the generalization ability and effectiveness of the proposed S2FL model in the case of three-modality data, i.e., HS, SAR, and DSM. Table \ref{tab:Au} details the classification results of different compared algorithms. Generally, there is an obvious improvement (around 4\%) in \textit{OA} when jointly using three modalities, compared to that using HS, SAR, and DSM independently. When the number of considered modalities increases to 3, the performance of those MA-based models dramatically declines, especially USMA that is a lack of label guidance. It is worth noted, however, that the CoSpace-based approaches modeled by either $\ell_{2}$-norm or $\ell_{1}$-norm, can be effectively extended to the case of three modalities, yielding significant improvement in classification accuracies. The feature selection works well in handling the issue of multiple heterogeneous data. That is, the classification result of the $\ell_{1}$-CoSpace is higher than that of $\ell_{2}$-CoSpace, which increases by 5.32\% points \textit{OA}. As excepted, the S2FL performs better than $\ell_{1}$-CoSpace with an increase of approximately 2 percentage points \textit{OA}. More importantly, the \textit{AA} value obtained by the S2FL is far higher than other competitors, where almost of each class can achieve the highest classification results. This, to a great extent, shows the superiority of our proposed shared and specific learning strategy (i.e., S2FL model).

Furthermore, the classification maps shown in Fig. \ref{fig:CM_Au_IF} also give a strong support to the aforementioned conclusion. The S2FL by the means of multiple modality data can obtain more realistic material identification in land cover mapping. In particular, the materials, e.g., \textit{Forest}, \textit{Residential Area}, \textit{Low Plants}, are classified in more smooth fashion, showing semantically meaningful structure.

\subsubsection{Ablation Study on the Use of Multiple Modalities}
To further verify the effectiveness and superiority of the proposed S2FL model in multimodal RS data feature learning and fusion, we will investigate the performance gain by using different combinations of multiple modalities. More specifically, three high-performance MFL algorithms, i.e., $\ell_{2}$-CoSpace, $\ell_{1}$-CoSpace, S2FL, are used for quantitative comparison on the HS-SAR-DSM \textit{Augsburg} datasets, as detailed in Table \ref{tab:Ablation}. On the whole, there are several important and intuitive conclusions, which can be summarized as follows:
\begin{itemize}
    \item The joint exploitation of multiple modalities can break the performance bottleneck in land cover classification. For example, the HS+SAR+DSM can usually obtain better classification results than the only use of two modalities. 
    \item Characterized by rich spectral information, the HS image tends to identify the materials at a more accurate level compared to SAR and DSM. 
    \item The HS+SAR results obtained by $\ell_{2}$-CoSpace are even slightly better than those of HS+SAR+DSM. This indicates that the $\ell_{2}$-CoSpace method fails to better fuse the multimodal information to some extent when the number of modalities increases.
    \item Feature selection guided by sparsity-promoting $\ell_{1}$-norm is an effective strategy for MFL. The resulting $\ell_{1}$-CoSpace observably outperforms $\ell_{2}$-CoSpace in different modality combinations.
    \item By decoupling the multimodal data into shared and specific components, S2FL is capable of better learning feature representations of multimodal data (\textit{cf.} $\ell_{2}$-CoSpace and $\ell_{1}$-CoSpace), further yielding higher classification performance in either two modalities or three modalities.
\end{itemize}

\begin{table*}[!t]
\centering
\caption{Ablation study for different modality combinations by using three high-performance MFL algorithms (i.e., $\ell_{2}$-CoSpace, $\ell_{1}$-CoSpace, S2FL). The best one is shown in bold.}
\vspace{2mm}
\resizebox{0.55\textwidth}{!}{ 
\begin{tabular}{c|c||ccc}
\toprule[1.5pt] Modality Combination & Methods & OA (\%)  & AA (\%)  & $\kappa$\\
\hline \hline
\multirow{3}{*}{SAR+DSM} & $\ell_{2}$-CoSpace & 68.14 & 39.29 & 0.5490\\
& $\ell_{1}$-CoSpace & 69.32 & 39.50 & 0.5649\\
& S2FL & \bf 69.49 & \bf 40.15 & \bf 0.5674 \\
\hline
\multirow{3}{*}{HS+DSM} & $\ell_{2}$-CoSpace & 70.79 & 53.41 & 0.5887\\
& $\ell_{1}$-CoSpace & \bf 81.17 & 52.75 & 0.7268\\
& S2FL & 81.11 & \bf 59.09 & \bf 0.7313 \\
\hline
\multirow{3}{*}{HS+SAR} & $\ell_{2}$-CoSpace & 76.82 & 54.55 & 0.6722\\
& $\ell_{1}$-CoSpace & 80.96 & 51.26 & 0.7243\\
& S2FL & \bf 82.34 & \bf 59.92 & \bf 0.7493 \\
\hline
\multirow{3}{*}{HS+SAR+DSM} & $\ell_{2}$-CoSpace & 76.17 & 56.13 & 0.6643\\
& $\ell_{1}$-CoSpace & 81.49 & 53.56 & 0.7327\\
& S2FL & \bf 83.36 & \bf 61.38 & \bf 0.7626 \\
\bottomrule[1.5pt]
\end{tabular}
}
\label{tab:Ablation}
\end{table*}

It is worth noting that currently-developed DL-based approaches have shown great potential in the fusion and representation learning of multimodal RS data, yet these methods inevitably suffer from various possible performance degradation. However, facing these problems, the proposed S2FL model could, to a great extent, offer capabilities that DL methods do not provide in the aspects of robustness, interpretability, and sensitivity to the training set size.

\section{Conclusion}
Land cover classification has long been considered as a main research topic in the RS and geoscience community. As a crucial step, feature extraction has been paid much attention by researchers. However, the feature representation ability extracted from only single RS data resources remains limited. Fortunately, the rapid development of RS imaging techniques makes multimodal RS data available on a large scale. To speed up the development of multimodal RS data processing and analysis, we in this paper aim at opening three multimodal RS benchmark datasets, they are homogeneous HS-MS \textit{Houston2013}, heterogeneous HS-SAR \textit{Berlin}, and three-modality HS-SAR-DSM \textit{Augsburg} datasets. Further, we also propose a novel MFL model, called S2FL, yielding more discriminative and compact feature blending by learning shared-modality and specific-modality representations. By comparing with previously-proposed advanced MFL methods, the S2FL model obtains the best classification performance on the three datasets, which is obviously superior to other competitors. We will open the three potential benchmark datasets and the MFL toolbox including newly-proposed S2FL model, contributing to the RS and information fusion community. In future work, we would like to further extend these datasets to a larger scale and also develop the corresponding feature learning models, e.g., based on more powerful deep learning techniques by embedding more interpretable knowledge or priors to guide the network optimization in the multimodal feature learning task.

\section*{Acknowledgement}
The authors would like to the Hyperspectral Image Analysis group at the University of Houston and the IEEE GRSS DFC2013 for providing the University of Houston HS dataset.

This work is jointly supported by the German Research Foundation (DFG) under grant ZH 498/7-2, by the European Research Council (ERC) under the European Union's Horizon 2020 research and innovation programme (grant agreement No. [ERC-2016-StG-714087], Acronym: \textit{So2Sat}), by the Helmholtz Association through the Framework of Helmholtz AI (grant  number:  ZT-I-PF-5-01) - Local Unit ``Munich Unit @Aeronautics, Space and Transport (MASTr)'' and Helmholtz Excellent Professorship ``Data Science in Earth Observation - Big Data Fusion for Urban Research''(W2-W3-100) and by the German Federal Ministry of Education and Research (BMBF) in the framework of the international future AI lab "AI4EO -- Artificial Intelligence for Earth Observation: Reasoning, Uncertainties, Ethics and Beyond" (Grant number: 01DD20001), by the National Natural Science Foundation of China (NSFC) under grant contracts No.41820104006. This work of J. Chanussot is also supported by the MIAI@Grenoble Alpes (ANR-19-P3IA-0003) and the AXA Research Fund.

\bibliographystyle{elsarticle-harv}
\bibliography{mybibfile}

\begin{thebibliography}{57}
\expandafter\ifx\csname natexlab\endcsname\relax\def\natexlab#1{#1}\fi
\expandafter\ifx\csname url\endcsname\relax
  \def\url#1{\texttt{#1}}\fi
\expandafter\ifx\csname urlprefix\endcsname\relax\def\urlprefix{URL }\fi

\bibitem[{Amici et~al.(2013)Amici, Piscini, Buongiorno, and
  Pieri}]{amici2013geological}
Amici, S., Piscini, A., Buongiorno, M.~F., Pieri, D., 2013. Geological
  classification of volcano teide by hyperspectral and multispectral satellite
  data. Int. J. Remote Sens. 34~(9-10), 3356--3375.

\bibitem[{Baumgartner et~al.(2012)Baumgartner, Gege, K{\"o}hler, Lenhard, and
  Schwarzmaier}]{baumgartner2012characterisation}
Baumgartner, A., Gege, P., K{\"o}hler, C., Lenhard, K., Schwarzmaier, T., 2012.
  Characterisation methods for the hyperspectral sensor hyspex at dlr's
  calibration home base. In: Sensors, Systems, and Next-Generation Satellites
  XVI. Vol. 8533. International Society for Optics and Photonics, p. 85331H.

\bibitem[{Bazaraa et~al.(2013)Bazaraa, Sherali, and
  Shetty}]{bazaraa2013nonlinear}
Bazaraa, M.~S., Sherali, H.~D., Shetty, C.~M., 2013. Nonlinear programming:
  theory and algorithms. John Wiley \& Sons.

\bibitem[{Bishop et~al.(2011)Bishop, Liu, and Mason}]{bishop2011hyperspectral}
Bishop, C.~A., Liu, J.~G., Mason, P.~J., 2011. Hyperspectral remote sensing for
  mineral exploration in pulang, yunnan province, china. Int. J. Remote Sens.
  32~(9), 2409--2426.

\bibitem[{Boyd et~al.(2011)Boyd, Parikh, and Chu}]{boyd2011distributed}
Boyd, S., Parikh, N., Chu, E., 2011. Distributed optimization and statistical
  learning via the alternating direction method of multipliers. Now Publishers
  Inc.

\bibitem[{Chen et~al.(2016)Chen, He, Ye, and Yuan}]{chen2016direct}
Chen, C., He, B., Ye, Y., Yuan, X., 2016. The direct extension of admm for
  multi-block convex minimization problems is not necessarily convergent. Math.
  Program. 155~(1-2), 57--79.

\bibitem[{Chen et~al.(2017)Chen, Li, Ghamisi, Jia, and Gu}]{chen2017deep}
Chen, Y., Li, C., Ghamisi, P., Jia, X., Gu, Y., 2017. Deep fusion of remote
  sensing data for accurate classification. IEEE Geosci. Remote Sens. Lett.
  14~(8), 1253--1257.

\bibitem[{Dalla~Mura et~al.(2015)Dalla~Mura, Prasad, Pacifici, Gamba,
  Chanussot, and Benediktsson}]{dalla2015challenges}
Dalla~Mura, M., Prasad, S., Pacifici, F., Gamba, P., Chanussot, J.,
  Benediktsson, J.~A., 2015. Challenges and opportunities of multimodality and
  data fusion in remote sensing. Proc. IEEE 103~(9), 1585--1601.

\bibitem[{Deng et~al.(2017)Deng, Lai, Peng, and Yin}]{deng2017parallel}
Deng, W., Lai, M.-J., Peng, Z., Yin, W., 2017. Parallel multi-block admm with o
  (1/k) convergence. J. Sci. Comput. 71~(2), 712--736.

\bibitem[{Ehlers et~al.(2010)Ehlers, Klonus, Johan~{\AA}strand, and
  Rosso}]{ehlers2010multi}
Ehlers, M., Klonus, S., Johan~{\AA}strand, P., Rosso, P., 2010. Multi-sensor
  image fusion for pansharpening in remote sensing. Int. J. Image Data Fusion
  1~(1), 25--45.

\bibitem[{Fang et~al.(2017)Fang, He, Li, Ghamisi, and
  Benediktsson}]{fang2017extinction}
Fang, L., He, N., Li, S., Ghamisi, P., Benediktsson, J.~A., 2017. Extinction
  profiles fusion for hyperspectral images classification. IEEE Trans. Geosci.
  Remote Sens. 56~(3), 1803--1815.

\bibitem[{Fauvel et~al.(2008)Fauvel, Benediktsson, Chanussot, and
  Sveinsson}]{fauvel2008spectral}
Fauvel, M., Benediktsson, J.~A., Chanussot, J., Sveinsson, J.~R., 2008.
  Spectral and spatial classification of hyperspectral data using svms and
  morphological profiles. IEEE Trans. Geosci. Remote Sens. 46~(11), 3804--3814.

\bibitem[{Gao et~al.(2020)Gao, Yuan, Li, Zhang, and Su}]{gao2020cloud}
Gao, J., Yuan, Q., Li, J., Zhang, H., Su, X., 2020. Cloud removal with fusion
  of high resolution optical and sar images using generative adversarial
  networks. Remote Sens. 12~(1), 191.

\bibitem[{Haklay and Weber(2008)}]{haklay2008openstreetmap}
Haklay, M., Weber, P., 2008. Openstreetmap: User-generated street maps. IEEE
  Pervas. Comput. 7~(4), 12--18.

\bibitem[{Hang et~al.(2020)Hang, Li, Ghamisi, Hong, Xia, and
  Liu}]{hang2020classification}
Hang, R., Li, Z., Ghamisi, P., Hong, D., Xia, G., Liu, Q., 2020. Classification
  of hyperspectral and lidar data using coupled cnns. IEEE Trans. Geosci.
  Remote Sens. 58~(7), 4939--4950.

\bibitem[{He and Niyogi(2004)}]{he2004locality}
He, X., Niyogi, P., 2004. Locality preserving projections. In: Proc. NIPS. pp.
  153--160.

\bibitem[{Heiden et~al.(2012)Heiden, Heldens, Roessner, Segl, Esch, and
  Mueller}]{heiden2012urban}
Heiden, U., Heldens, W., Roessner, S., Segl, K., Esch, T., Mueller, A., 2012.
  Urban structure type characterization using hyperspectral remote sensing and
  height information. Landsc. Urban Plan. 105~(4), 361--375.

\bibitem[{Hong et~al.(2020{\natexlab{a}})Hong, Chanussot, Yokoya, Kang, and
  Zhu}]{hong2020learning}
Hong, D., Chanussot, J., Yokoya, N., Kang, J., Zhu, X.~X., 2020{\natexlab{a}}.
  Learning-shared cross-modality representation using multispectral-lidar and
  hyperspectral data. IEEE Geosci. Remote Sens. Lett. 17~(8), 1470--1474.

\bibitem[{Hong et~al.(2020{\natexlab{b}})Hong, Gao, Yao, Zhang, Antonio, and
  Chanussot}]{hong2020graph}
Hong, D., Gao, L., Yao, J., Zhang, B., Antonio, P., Chanussot, J.,
  2020{\natexlab{b}}. Graph convolutional networks for hyperspectral image
  classification. IEEE Trans. Geosci. Remote Sens.DOI:
  10.1109/TGRS.2020.3015157.

\bibitem[{Hong et~al.(2021)Hong, Gao, Yokoya, Yao, Chanussot, Du, and
  Zhang}]{hong2021more}
Hong, D., Gao, L., Yokoya, N., Yao, J., Chanussot, J., Du, Q., Zhang, B., 2021.
  More diverse means better: Multimodal deep learning meets remote-sensing
  imagery classification. IEEE Trans. Geosci. Remote Sens. 59~(5), 4340--4354.

\bibitem[{Hong et~al.(2019{\natexlab{a}})Hong, Yokoya, Chanussot, and
  Zhu}]{hong2019cospace}
Hong, D., Yokoya, N., Chanussot, J., Zhu, X., 2019{\natexlab{a}}. Co{S}pace:
  Common subspace learning from hyperspectral-multispectral correspondences.
  IEEE Trans. Geosci. Remote Sens. 57~(7), 4349--4359.

\bibitem[{Hong et~al.(2019{\natexlab{b}})Hong, Yokoya, Chanussot, and
  Zhu}]{hong2019augmented}
Hong, D., Yokoya, N., Chanussot, J., Zhu, X.~X., 2019{\natexlab{b}}. An
  augmented linear mixing model to address spectral variability for
  hyperspectral unmixing. IEEE Trans. Image Process. 28~(4), 1923--1938.

\bibitem[{Hong et~al.(2019{\natexlab{c}})Hong, Yokoya, Ge, Chanussot, and
  Zhu}]{hong2019learnable}
Hong, D., Yokoya, N., Ge, N., Chanussot, J., Zhu, X.~X., 2019{\natexlab{c}}.
  Learnable manifold alignment (lema): A semi-supervised cross-modality
  learning framework for land cover and land use classification. ISPRS J.
  Photogramm. Remote Sens. 147, 193--205.

\bibitem[{Hu et~al.(2019)Hu, Hong, and Zhu}]{hu2019mima}
Hu, J., Hong, D., Zhu, X.~X., 2019. Mima: Mapper-induced manifold alignment for
  semi-supervised fusion of optical image and polarimetric sar data. IEEE
  Trans. Geosci. Remote Sens. 57~(11), 9025--9040.

\bibitem[{Kurz et~al.(2011)Kurz, Rosenbaum, Leitloff, Meynberg, and
  Reinartz}]{kurz2011real}
Kurz, F., Rosenbaum, D., Leitloff, J., Meynberg, O., Reinartz, P., 2011. Real
  time camera system for disaster and traffic monitoring. In: Proc. SMPR. pp.
  1--6.

\bibitem[{Lai and Osher(2014)}]{lai2014splitting}
Lai, R., Osher, S., 2014. A splitting method for orthogonality constrained
  problems. J. Sci. Comput. 58~(2), 431--449.

\bibitem[{Liao et~al.(2016)Liao, Qian, Zhou, and Tang}]{liao2016manifold}
Liao, D., Qian, Y., Zhou, J., Tang, Y.~Y., 2016. A manifold alignment approach
  for hyperspectral image visualization with natural color. IEEE Trans. Geosci.
  Remote Sens. 54~(6), 3151--3162.

\bibitem[{Liao et~al.(2014)Liao, Pi{\v{z}}urica, Bellens, Gautama, and
  Philips}]{liao2014generalized}
Liao, W., Pi{\v{z}}urica, A., Bellens, R., Gautama, S., Philips, W., 2014.
  Generalized graph-based fusion of hyperspectral and lidar data using
  morphological features. IEEE Geosci. Remote Sens. Lett. 12~(3), 552--556.

\bibitem[{Lin et~al.(2010)Lin, Chen, and Ma}]{lin2010augmented}
Lin, Z., Chen, M., Ma, Y., 2010. The augmented lagrange multiplier method for
  exact recovery of corrupted low-rank matrices. arXiv preprint
  arXiv:1009.5055.

\bibitem[{Liu et~al.(2017)Liu, Du, Tong, Samat, Pan, and Ma}]{liu2017band}
Liu, S., Du, Q., Tong, X., Samat, A., Pan, H., Ma, X., 2017. Band
  selection-based dimensionality reduction for change detection in
  multi-temporal hyperspectral images. Remote Sens. 9~(10), 1008.

\bibitem[{Liu et~al.(2019)Liu, Marinelli, Bruzzone, and Bovolo}]{liu2019review}
Liu, S., Marinelli, D., Bruzzone, L., Bovolo, F., 2019. A review of change
  detection in multitemporal hyperspectral images: Current techniques,
  applications, and challenges. IEEE Geosci. Remote Sens. Mag. 7~(2), 140--158.

\bibitem[{Liu et~al.(2020)Liu, Zheng, Dalponte, and Tong}]{liu2020novel}
Liu, S., Zheng, Y., Dalponte, M., Tong, X., 2020. A novel fire index-based
  burned area change detection approach using landsat-8 oli data. Eur. J.
  Remote Sens. 53~(1), 104--112.

\bibitem[{Ma et~al.(2018)Ma, Tong, Liu, Li, and Ma}]{ma2018multisource}
Ma, X., Tong, X., Liu, S., Li, C., Ma, Z., 2018. A multisource remotely sensed
  data oriented method for “ghost city” phenomenon identification. IEEE J.
  Sel. Top. Appl. Earth Obs. Remote Sens. 11~(7), 2310--2319.

\bibitem[{Mart{\'\i}nez and Kak(2001)}]{martinez2001pca}
Mart{\'\i}nez, A.~M., Kak, A.~C., 2001. Pca versus lda. IEEE Trans. Pattern
  Anal. Mach. Intell. 23~(2), 228--233.

\bibitem[{Nativi et~al.(2015)Nativi, Mazzetti, Santoro, Papeschi, Craglia, and
  Ochiai}]{nativi2015big}
Nativi, S., Mazzetti, P., Santoro, M., Papeschi, F., Craglia, M., Ochiai, O.,
  2015. Big data challenges in building the global earth observation system of
  systems. Environ. Model. Softw. 68, 1--26.

\bibitem[{Ngiam et~al.(2011)Ngiam, Khosla, Kim, Nam, Lee, and
  Ng}]{ngiam2011multimodal}
Ngiam, J., Khosla, A., Kim, M., Nam, J., Lee, H., Ng, A., 2011. Multimodal deep
  learning. In: Proc. ICML. pp. 689--696.

\bibitem[{Okujeni et~al.(2016)Okujeni, van~der Linden, and
  Hostert}]{okujeni2016berlin}
Okujeni, A., van~der Linden, S., Hostert, P., 2016. Berlin-urban-gradient
  dataset 2009-an enmap preparatory flight campaign.

\bibitem[{Pournemat et~al.(2020)Pournemat, Adibi, and
  Chanussot}]{pournemat111semisupervised}
Pournemat, A., Adibi, P., Chanussot, J., 2020. Semisupervised charting for
  spectral multimodal manifold learning and alignment. Pattern Recognit. 111,
  107645.

\bibitem[{Rasti et~al.(2017)Rasti, Ghamisi, and
  Gloaguen}]{rasti2017hyperspectral}
Rasti, B., Ghamisi, P., Gloaguen, R., 2017. Hyperspectral and lidar fusion
  using extinction profiles and total variation component analysis. IEEE Trans.
  Geosci. Remote Sens. 55~(7), 3997--4007.

\bibitem[{Rasti et~al.(2020)Rasti, Hong, Hang, Ghamisi, Kang, Chanussot, and
  Benediktsson}]{rasti2020feature}
Rasti, B., Hong, D., Hang, R., Ghamisi, P., Kang, X., Chanussot, J.,
  Benediktsson, J., 2020. Feature extraction for hyperspectral imagery: The
  evolution from shallow to deep: Overview and toolbox. IEEE Geosci. Remote
  Sens. Mag. 8~(4), 60--88.

\bibitem[{Schmitt and Zhu(2016)}]{schmitt2016}
Schmitt, M., Zhu, X.~X., 2016. Data fusion and remote sensing: An ever-growing
  relationship. IEEE Geoscience and Remote Sensing Magazine 4~(4), 6--23.

\bibitem[{Siebels et~al.(2020)Siebels, Go{\"\i}ta, and
  Germain}]{siebels2020estimation}
Siebels, K., Go{\"\i}ta, K., Germain, M., 2020. Estimation of mineral abundance
  from hyperspectral data using a new supervised neighbor-band ratio unmixing
  approach. IEEE Trans. Geosci. Remote Sens. 58~(10), 6754--6766.

\bibitem[{Tuia and Camps-Valls(2016)}]{tuia2016kernel}
Tuia, D., Camps-Valls, G., 2016. Kernel manifold alignment for domain
  adaptation. PLOS One 11~(2), e0148655.

\bibitem[{Tuia et~al.(2014)Tuia, Volpi, Trolliet, and
  Camps-Valls}]{tuia2014semisupervised}
Tuia, D., Volpi, M., Trolliet, M., Camps-Valls, G., 2014. Semisupervised
  manifold alignment of multimodal remote sensing images. IEEE Trans. Geosci.
  Remote Sens. 52~(12), 7708--7720.

\bibitem[{Van~der Meer et~al.(2012)Van~der Meer, Van~der Werff, Van~Ruitenbeek,
  Hecker, Bakker, Noomen, Van Der~Meijde, Carranza, De~Smeth, and
  Woldai}]{van2012multi}
Van~der Meer, F.~D., Van~der Werff, H.~M., Van~Ruitenbeek, F.~J., Hecker,
  C.~A., Bakker, W.~H., Noomen, M.~F., Van Der~Meijde, M., Carranza, E. J.~M.,
  De~Smeth, J.~B., Woldai, T., 2012. Multi-and hyperspectral geologic remote
  sensing: A review. Int. J. Appl. Earth Obs. Geoinf. 14~(1), 112--128.

\bibitem[{Wang and Mahadevan(2011)}]{wang2011heterogeneous}
Wang, C., Mahadevan, S., 2011. Heterogeneous domain adaptation using manifold
  alignment. In: Proc. IJCAI. Vol.~22. p. 1541.

\bibitem[{Wang et~al.(2018)Wang, Cao, and Xu}]{wang2018convergence}
Wang, F., Cao, W., Xu, Z., 2018. Convergence of multi-block bregman admm for
  nonconvex composite problems. Sci. China Inform. Sci. 61~(12), 122101.

\bibitem[{Wei et~al.(2015)Wei, Bioucas-Dias, Dobigeon, and
  Tourneret}]{wei2015hyperspectral}
Wei, Q., Bioucas-Dias, J., Dobigeon, N., Tourneret, J.-Y., 2015. Hyperspectral
  and multispectral image fusion based on a sparse representation. IEEE Trans.
  Geosci. Remote Sens. 53~(7), 3658--3668.

\bibitem[{Weng(2009)}]{weng2009thermal}
Weng, Q., 2009. Thermal infrared remote sensing for urban climate and
  environmental studies: Methods, applications, and trends. ISPRS J.
  Photogramm. Remote Sens. 64~(4), 335--344.

\bibitem[{Xia et~al.(2019)Xia, Liao, and Du}]{xia2019hyperspectral}
Xia, J., Liao, W., Du, P., 2019. Hyperspectral and lidar classification with
  semisupervised graph fusion. IEEE Geosci. Remote Sens. Lett. 17~(4),
  666--670.

\bibitem[{Xie and Weng(2017)}]{xie2017spatiotemporally}
Xie, Y., Weng, Q., 2017. Spatiotemporally enhancing time-series dmsp/ols
  nighttime light imagery for assessing large-scale urban dynamics. ISPRS J.
  Photogramm. Remote Sens. 128, 1--15.

\bibitem[{Xu et~al.(2019)Xu, Wu, Chanussot, Comon, and Wei}]{xu2019nonlocal}
Xu, Y., Wu, Z., Chanussot, J., Comon, P., Wei, Z., 2019. Nonlocal coupled
  tensor cp decomposition for hyperspectral and multispectral image fusion.
  IEEE Trans. Geosci. Remote Sens. 58~(1), 348--362.

\bibitem[{Yao et~al.(2019)Yao, Meng, Zhao, Cao, and Xu}]{yao2019nonconvex}
Yao, J., Meng, D., Zhao, Q., Cao, W., Xu, Z., 2019. Nonconvex-sparsity and
  nonlocal-smoothness-based blind hyperspectral unmixing. IEEE Trans. Image
  Process. 28~(6), 2991--3006.

\bibitem[{Yokoya et~al.(2017)Yokoya, Ghamisi, and Xia}]{yokoya2017multimodal}
Yokoya, N., Ghamisi, P., Xia, J., 2017. Multimodal, multitemporal, and
  multisource global data fusion for local climate zones classification based
  on ensemble learning. In: Proc. IGARSS. IEEE, pp. 1197--1200.

\bibitem[{Zhou et~al.(2016)Zhou, Zhang, and Lin}]{zhou2016bilevel}
Zhou, P., Zhang, C., Lin, Z., 2016. Bilevel model-based discriminative
  dictionary learning for recognition. IEEE Trans. Image Process. 26~(3),
  1173--1187.

\bibitem[{Zhu et~al.(2019)Zhu, Hou, Weng, and Chen}]{zhu2019integrating}
Zhu, X., Hou, Y., Weng, Q., Chen, L., 2019. Integrating uav optical imagery and
  lidar data for assessing the spatial relationship between mangrove and
  inundation across a subtropical estuarine wetland. ISPRS J. Photogramm.
  Remote Sens. 149, 146--156.

\bibitem[{Zhu et~al.(2017)Zhu, Tuia, Mou, Xia, Zhang, Xu, and
  Fraundorfer}]{zhu2017}
Zhu, X.~X., Tuia, D., Mou, L., Xia, G.-S., Zhang, L., Xu, F., Fraundorfer, F.,
  2017. Deep learning in remote sensing: A comprehensive review and list of
  resources. IEEE Geoscience and Remote Sensing Magazine 5~(4), 8--36.

\end{thebibliography}
\end{document}